%% file: main.tex
\documentclass{article}

\PassOptionsToPackage{numbers, compress}{natbib}
\usepackage[final,main]{neurips_2026}

\usepackage[utf8]{inputenc} 
\usepackage[T1]{fontenc}    
\usepackage[hidelinks]{hyperref}       
\usepackage{url}            
\usepackage{booktabs}       
\usepackage{amsfonts}       
\usepackage{amsmath}        
\usepackage{amssymb}        
\usepackage{amsthm}         
\usepackage{nicefrac}       
\usepackage{microtype}      
\usepackage{xcolor}         
\usepackage{graphicx}       
\usepackage{algorithm}      
\usepackage{algorithmic}    
\usepackage{multirow}       
\usepackage{subcaption}     
\usepackage{float}          
\usepackage{tikz}           
\usetikzlibrary{arrows.meta, positioning, calc, fit, backgrounds, decorations.pathreplacing}
\usepackage{placeins}       

\newtheorem{assumption}{Assumption}
\newtheorem{proposition}{Proposition}

\newcommand{\loss}{\mathcal{L}}
\newcommand{\encoder}{f}
\newcommand{\probe}{h}

\newcommand{\sg}{\operatorname{sg}}

\title{Probe-Guided Gradient Balancing 

for Multimodal Learning}

\author{%
  Vu Vo \\
  School of Science, Engineering \& Technology \\
  RMIT University \\
  \texttt{s3575819@rmit.edu.vn} \\
  \And
  Haytham M. Fayek \\
  School of Computing Technologies \\
  RMIT University \\
  \texttt{haytham.fayek@ieee.org} \\
  \And
  Thuy Nguyen \\
  School of Science, Engineering \& Technology \\
  RMIT University \\
  \texttt{thuy.nguyen43@rmit.edu.vn} \\
}

\begin{document}

\maketitle

\begin{abstract}

Jointly trained multimodal networks frequently underutilize their weaker modalities, as faster-learning modalities dominate the shared fusion objective and suppress the gradient signal reaching slower ones. Most existing approaches throttle the dominant modality, whereas directly boosting the weaker modality often fails because monitoring signals derived from the joint loss are entangled with, and thus dominated by, the stronger modality.
We propose a new gradient modulation approach, probe-guided gradient boosting (PGGB), to address this problem by probing and boosting the weak modalities while preserving the strength of the dominant modality. 
Probing is performed by attaching a lightweight linear classifier to each modality’s stop-gradient features to estimate the corresponding representation quality score. The unbiased gap between these scores across modalities, the \emph{utilization gap}, serves as an imbalance signal that defines a bounded and smoothed scaling factor adaptively reweighing the gradients of weaker modalities, enabling stable and effective rebalancing during joint optimization.
PGGB is composable with throttling methods as the intervention leaves the loss and fusion architecture unchanged. 
Extensive experiments were conducted across eight benchmarks spanning four domains and 2–4 modalities. The results show that PGGB outperforms state-of-the-art approaches on highly imbalanced multimodal datasets, while remaining competitive on benchmarks with low or no modality imbalance.
Bounded scaling, self-attenuation, and a standard-SGD descent bound were established under standard smoothness and bounded-variance assumptions. 
Code is available at \url{https://github.com/VuVoCodes/PGGB}.
\end{abstract}

\section{Introduction}
\label{sec:intro}



Multimodal learning integrates complementary signals across heterogeneous sources, including audio, visual, and textual streams among others~\citep{baltrusaitis2019multimodal}. Under mild assumptions on latent structure, multimodal models are provably superior to their unimodal counterparts~\citep{huang2021multimodal}. Despite this theoretical promise, jointly trained multimodal networks frequently fail to fully exploit all available modalities, often performing no better than their strongest unimodal component. This \emph{modality laziness} effect~\citep{wang2020gblending,peng2022ogmge,du2023suppression} has been observed for instance in audio-dominant emotion recognition on the CREMA-D~\citep{cao2014cremad} dataset.


Existing methods mostly focusing on \emph{throttling-only}, intervening primarily by damping the dominant modality's gradient~\citep{peng2022ogmge,wei2024mmpareto,guo2024cggm,wei2024opm,fan2023pmr,kontras2024mlgm}. Some include nominally bidirectional terms~\citep{li2023agm,gao2025arm,huang2025inforeg,wei2025improving}, but throttling empirically dominates, and \citet{wei2024opm}'s symmetric boost-direction variant fails to close the modality gap. In late-fusion, every encoder's gradient flows through a shared classifier whose loss-derivative is dominated by the strong modality's contribution to the fusion loss. Amplifying the weak encoder's gradient therefore rescales an already-corrupted shared signal. Recent methods take partial steps~\citep{yang2025iprm,jiang2025aug} but continue to monitor imbalance through signals coupled to the joint training objective.


We propose \emph{probe-guided gradient boosting} (PGGB): lightweight linear probes~\citep{alain2017understanding} on stop-gradient features\footnote{The stop-gradient operator passes activations forward but blocks the gradient in the backward pass~\citep{vandenoord2017vqvae}.} yield a per-modality representation-quality estimate, whose cross-modality spread defines the \emph{utilization gap}, independent of the shared classifier's loss-derivative, removing the chain-rule coupling that impaired prior boost-the-weak attempts~\citep{wei2024opm}. The signal drives a bounded multiplicative factor on the weaker encoder's gradient, smoothed by an exponential moving average (EMA)~\citep{tarvainen2017mean}. Because the factor acts only on encoder gradients and modifies neither the loss nor the fusion architecture, it is \emph{composable} with throttling methods, multiplying with their suppression factors to produce two-sided pressure toward balance.


We evaluate PGGB on eight benchmarks spanning audio-visual, sentiment, text-image, and medical-segmentation tasks (details in Section~\ref{sec:datasets}). The main gain is observed on the high-imbalance CREMA-D benchmark (3-frame), where composing PGGB with a representative gradient-modulation baseline~\citep{peng2022ogmge} reaches $71.45 \pm 1.71\%$ accuracy, $+2.31$~pp over the same baseline alone and $+9.86$~pp over joint training. On the four low-imbalance benchmarks (AVE, Kinetics-Sounds, Twitter15, Sarcasm), PGGB alone is the best-performing method on all four (Table~\ref{tab:main_results}). On the high-imbalance text-dominant sentiment benchmarks (CMU-MOSI, CH-SIMS), where text alone is near-multimodally sufficient and prior balancing methods report only modest gains~\citep{liu2022chsimsv2,wei2025dgl}, PGGB+OGM-GE remains within seed standard deviation of the strongest baseline ($-0.08$ and $-0.04$~pp). The OGM-GE composition variant has two isolated regressions versus joint training (Kinetics-Sounds $-1.72$~pp, Sarcasm $-0.64$~pp) noted in Section~\ref{sec:conclusion}, while several gradient-modulation baselines degrade on multiple datasets (e.g., OGM-GE~\citep{peng2022ogmge} regresses on Kinetics-Sounds by $-1.80$~pp, Table~\ref{tab:main_results}).

\paragraph{Contributions.}
\begin{itemize}
    \item We identify a \emph{throttle-only bias} in the modality-balancing literature and show that its natural complement, boosting the weaker modality, fails when the monitoring signal is coupled to the joint training objective via shared loss derivatives in late fusion.

    \item We propose \emph{decoupled probe monitoring}, a diagnostic attaching linear probes to stop-gradient features, yielding an unbiased \emph{utilization gap} estimate (the spread of per-modality probe accuracies) independent of the joint objective, and this decoupling enables weak-modality boosting. We introduce \emph{PGGB}, one instantiation converting the monitoring signal into a bounded multiplicative scaling on the weaker encoder's gradients, composable with existing throttling methods (Appendix~\ref{app:composability} validates non-destructive composition with four such methods).

    \item We validate \emph{PGGB} on eight benchmarks spanning audio-visual, sentiment, text-image, and medical-segmentation tasks. \emph{PGGB} composed with throttling outperforms state-of-the-art baselines on high-imbalance audio-visual (CREMA-D), matches the strongest available baseline on medical segmentation (BraTS~2021, $+0.28$~pp, within seed variance), and remains competitive within seed variance on the high-imbalance text-dominant sentiment benchmarks (CMU-MOSI, CH-SIMS), and \emph{PGGB} used alone is the best-performing method on all four low-imbalance datasets (AVE, Kinetics-Sounds, Twitter15, Sarcasm).
\end{itemize}

\section{Related work}
\label{sec:related}

\paragraph{Modality imbalance in joint training.}

Jointly trained multimodal networks frequently underperform their best unimodal component~\citep{wang2020gblending}. Under a Gaussian mixture model, \citet{huang2022modality} prove that joint training learns only a subset of modalities and derive bounds on the duration of the resulting unimodal phase, and \citet{du2023suppression} characterize the conditions under which individual modality features fail to be learned. \citet{zhang2024unimodal} further show, in deep linear networks, that this unimodal phase depends on architecture depth and initialization. Modality imbalance is therefore a structural property of joint optimization, rooted in the shared loss through which all encoder gradients flow.

\paragraph{Gradient modulation for modality balance.}

The dominant response has been to modulate gradient magnitudes during backpropagation. Multi-task precursors GradNorm~\citep{chen2018gradnorm} and PCGrad~\citep{yu2020pcgrad} inspired a line of multimodal work. OGM-GE~\citep{peng2022ogmge} dampens the dominant modality's gradient via a ratio of per-modality contributions to the target-class score, refined by AGM~\citep{li2023agm}, CGGM~\citep{guo2024cggm}, and MLGM~\citep{kontras2024mlgm}. Although some methods include bidirectional terms, throttling empirically dominates and boost-only variants remain ineffective: the symmetric boost direction (OGM*) reported by~\citet{wei2024opm} shows mixed performance in closing the modality gap, which we attribute to coupled monitoring signals affected by the dominant modality (Section~\ref{sec:method}). M-SAM~\citep{nowdeh2025msam} pairs Shapley attribution with sharpness-aware perturbation, inheriting the same coupled-monitoring limitation.

\paragraph{Alternative approaches to modality balance.}

Beyond gradient scaling, a broader set of interventions has been proposed. These include feed-forward feature modulation~\citep{wei2024opm} and Pareto-optimal aggregation~\citep{wei2024mmpareto}. Modality-alternating schemes interleave unimodal and joint updates~\citep{hua2024reconboost,zhang2024mla}. Diagnostic methods detect imbalance through separability~\citep{wei2024dr}, Fisher information~\citep{huang2025inforeg}, or bias-variance decomposition~\citep{wei2025improving}. Other approaches intervene at the data level~\citep{ma2025remixing,yu2026tcmax}. Per-modality learning-rate methods~\citep{yao2022mslr,guerra2025miles} are an adjacent class operating on the optimizer step rather than the gradient. Despite their diversity, these methods share a common design choice. Imbalance is detected through signals \emph{coupled} to the joint objective, such as gradient ratios, per-modality losses, or live-model predictions (with the partial exception of PMR's decoupled unimodal prototypes~\citep{fan2023pmr}). These are precisely the quantities distorted by dominant-modality dynamics, which limits the reliability of the resulting corrections.

\paragraph{Representation monitoring and probing.}

Linear probes on frozen representations~\citep{alain2017understanding} are a standard tool for assessing representation quality. Several multimodal methods adopt probe-like monitoring but retain coupling to the joint objective. IPRM's two-pass forward strategy compares unimodal and fused predictions of the live model via KL divergence~\citep{yang2025iprm}, CGGM's auxiliary classifiers participate in the training loss~\citep{guo2024cggm}, and AUG~\citep{jiang2025aug} relies on loss-based weakness signals that inherit the dominant modality's bias. PMR~\citep{fan2023pmr} moves toward decoupling via unimodal prototypes but the prototype signal feeds a gradient coefficient on the shared encoder, while DI-MML~\citep{fan2024dimml} detaches only the unimodal pre-fusion stage. In contrast, we train \emph{fully decoupled} linear probes on stop-gradient features with separate optimizers and evaluate them on held-out within-batch samples to prevent probe overfitting. The resulting utilization gap drives a bounded multiplicative boost on the weaker encoder's gradients, so the imbalance estimate is not affected by the dynamics it aims to correct.

\section{Probe-Guided Gradient Boosting}
\label{sec:method}

\begin{figure}[t]
\centering
\includegraphics[width=0.95\textwidth]{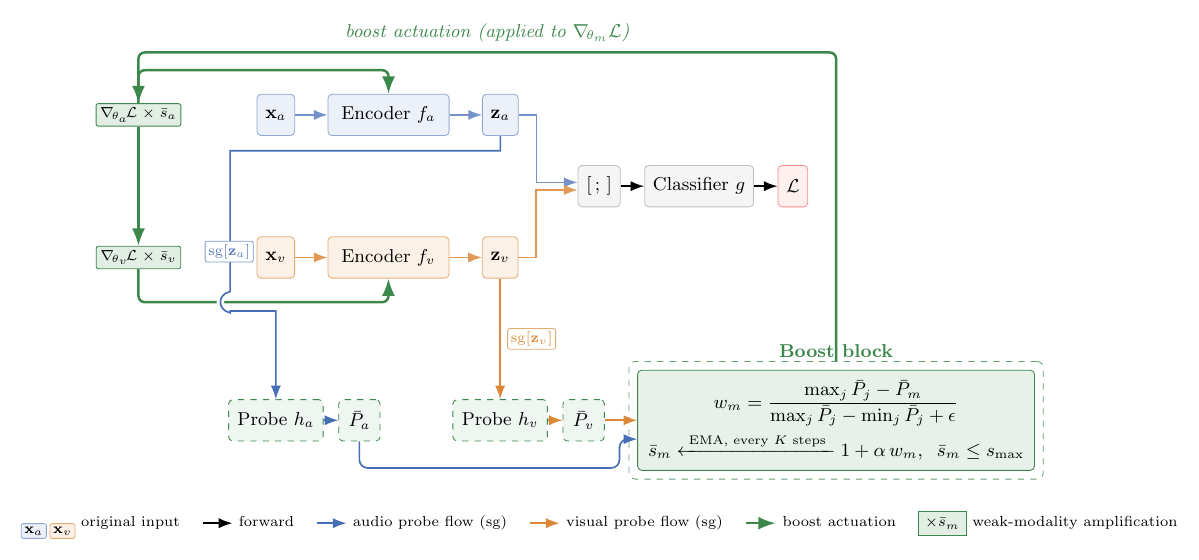}
\caption{An Illustration of probe-guided gradient boosting on 2 modalities. \textbf{Forward path (top):} features $\mathbf{z}_m$ are concatenated and classified via the shared fusion classifier $g$, producing $\loss$. \textbf{Probe monitoring (bottom):} stop-gradient features $\sg[\mathbf{z}_m]$ are evaluated by linear probes along solid per-modality probe-flow lines (audio in blue, visual in orange), yielding per-modality accuracies $\bar{P}_m$. The boost block computes the relative-gap weight $w_m$ and a smoothed scale $\bar{s}_m$ (EMA-smoothed, refreshed only at probe-evaluation events every $K$ steps and held constant in between), clipped at $s_{\max}$. The actuation rail (thick green) applies $\bar{s}_m$ to the per-modality gradient $\nabla_{\!\theta_m}\loss$. Probes never backpropagate into encoders, so the monitoring signal is structurally decoupled from the fusion objective.}
\label{fig:architecture}
\end{figure}

We present PGGB in four stages: problem formulation (\S\ref{sec:problem}), decoupled probe monitoring (\S\ref{sec:probes}), the gradient boosting mechanism (\S\ref{sec:boost}), and composability with existing balancing methods (\S\ref{sec:compose}). Figure~\ref{fig:architecture} depicts the data flow.

\subsection{Problem formulation}
\label{sec:problem}

Each multimodal sample comprises $M$ inputs $\{\mathbf{x}_m\}_{m=1}^{M}$, one per modality. Encoder $\encoder_m$ with parameters $\theta_m$ maps $\mathbf{x}_m$ to $\mathbf{z}_m$:
\begin{equation}
\label{eq:encoder}
\mathbf{z}_m = \encoder_m(\mathbf{x}_m;\,\theta_m) \in \mathbb{R}^d.
\end{equation}
Each $\mathbf{z}_m$ is a real value feature vector in $\mathbb{R}^d$, with $d$ common across modalities. 
Features are concatenated and mapped by the shared fusion classifier $g$ with parameters $\phi$ to a prediction $\hat{y} = g([\mathbf{z}_1;\ldots;\mathbf{z}_M];\,\phi)$, scored against the target $y$:
\begin{equation}
\label{eq:joint_loss}
\loss_{\text{fusion}} = \ell\!\bigl(\hat{y},\; y\bigr).
\end{equation}
Here $\ell$ is a differentiable per-sample loss mapping a prediction-target pair to a non-negative scalar. By the chain rule, the gradient with respect to encoder parameters decomposes as
\begin{equation}
\label{eq:gradient_chain}
{\partial \loss_{\text{fusion}}}/{\partial \theta_m} = ({\partial \loss_{\text{fusion}}}/{\partial g})\,({\partial g}/{\partial \mathbf{z}_m})\,({\partial \mathbf{z}_m}/{\partial \theta_m}).
\end{equation}
The term ${\partial \loss_{\text{fusion}}}/{\partial g}$ is shared across all modalities, since all encoder outputs flow through the same fusion head. This shared dependency is the root of coupled monitoring failure: when one modality (say $m{=}1$) converges faster and begins to dominate the head's outputs, ${\partial \loss_{\text{fusion}}}/{\partial g}$ increasingly reflects the dominant modality's contribution. Any monitoring signal derived from $\loss_{\text{fusion}}$ or its gradients, including the discrepancy ratio used by OGM-GE~\citep{peng2022ogmge} and the auxiliary classifier losses used by CGGM~\citep{guo2024cggm}, inherits this bias.

\FloatBarrier
\subsection{Decoupled probe monitoring}
\label{sec:probes}

We attach a lightweight linear probe classifier $\probe_m\colon \mathbb{R}^d \to \mathbb{R}^C$ (with $C$ outputs) to each encoder, parameterized by $\psi_m$. The resulting probe accuracies define an imbalance signal, the \emph{utilization gap} (Eq.~\eqref{eq:utilization_gap}), independent of the joint training objective. The probes are trained on stop-gradient features $\sg[\mathbf{z}_m]$, where $\sg[\cdot]$ denotes the stop-gradient operator~\citep{vandenoord2017vqvae} (identity at the forward pass, zero partial derivatives at the backward pass):
\begin{equation}
\label{eq:probe_loss}
\loss_{\text{probe},m} = \ell_{\text{probe}}\!\bigl(\probe_m(\sg[\mathbf{z}_m];\,\psi_m),\; y\bigr).
\end{equation}
Each probe has its own optimizer that updates only $\psi_m$, and no gradient flows into $\theta_m$. Since $\sg[\mathbf{z}_m]$ halts backward propagation at the feature level, $\loss_{\text{probe},m}$ depends only on the quality of $\mathbf{z}_m$ as a representation, not on how it contributes through $g$. This is precisely why decoupling matters: the probe accuracy $P_m$ reflects the representation quality of $\mathbf{z}_m$ alone, independent of the weight that the fusion classifier $g$ assigns to modality $m$.

\paragraph{Split-batch evaluation and on-the-fly estimation.}
Every $K$ training iterations, we partition the (shuffled) mini-batch into two disjoint halves, training probes on the first half and evaluating on the second. This yields an unbiased estimate $P_m$ of the current probe's training-distribution accuracy (conditioned on $\psi_m^{(t)}$), the evaluation half is held out from probe training and drawn i.i.d.\ from the training distribution. To convert this per-evaluation sample into a stable estimator, we accumulate an exponential moving average (EMA):
\begin{equation}
\label{eq:ema_probe}
\bar{P}_m \leftarrow \beta \cdot P_m + (1 - \beta) \cdot \bar{P}_m,
\end{equation}
where $\beta \in [0,1]$ is the smoothing coefficient. The EMA acts as a first-order low-pass filter on the probe signal, attenuating batch-to-batch fluctuation while tracking sustained changes in utilization. At $\beta = 0.1$ (\S\ref{sec:datasets}), the steady-state variance is $\beta/(2-\beta)\,\sigma^{2} \approx 0.053\,\sigma^{2}$ under an i.i.d.\ assumption with per-sample variance $\sigma^{2}$, approximately $19\times$ smaller than a single-batch estimate. The \emph{utilization gap} is the spread of smoothed probe accuracies:
\begin{equation}
\label{eq:utilization_gap}
\delta = \max_{m}\,\bar{P}_m - \min_{m}\,\bar{P}_m.
\end{equation}
Appendix~\ref{app:probe_stability} verifies empirically that the EMA estimator substantially reduces mini-batch noise in the probe signal during training.

\subsection{Probe-guided gradient boosting}
\label{sec:boost}

Given the EMA-smoothed probe accuracies $\{\bar{P}_m\}_{m=1}^{M}$ (Eq.~\eqref{eq:ema_probe}), we compute a per-modality boost scale that amplifies the gradients of weaker encoders (full pseudocode in Appendix~\ref{app:algorithm}, Algorithm~\ref{alg:pgb}). The procedure involves three steps.

\paragraph{Step 1: Relative weakness.}
For each modality $m$, we compute a normalized weakness score $w_m \in [0,1]$ (zero for the strongest modality, one for the weakest):
\begin{equation}
\label{eq:weakness}
w_m = \frac{\max_j\,\bar{P}_j - \bar{P}_m}{\max_j\,\bar{P}_j - \min_j\,\bar{P}_j + \epsilon},
\end{equation}
where $j$ ranges over all modalities $\{1, \ldots, M\}$, and $\epsilon$ is a small positive constant serving as a numerical guard against division by zero.

\paragraph{Step 2: Boost scale.}
$w_m$ is converted to a multiplicative scaling factor:
\begin{equation}
\label{eq:boost_scale}
s_m = \min\!\bigl(1 + \alpha \cdot w_m,\; s_{\max}\bigr),
\end{equation}
where $\alpha > 0$ controls the boost strength and $s_{\max} > 1$ caps the maximum amplification to prevent extreme scaling. When modalities are balanced ($\delta$ small), weakness scores become near zero and $s_m \approx 1$, so the method applies no intervention (Proposition~\ref{prop:attenuation} gives the precise continuity bound).
At exact balance all $\bar{P}_j$ coincide, so the numerator of Eq.~\eqref{eq:weakness} vanishes, the denominator equals $\epsilon > 0$, and $s_m = 1$ exactly.

\paragraph{Step 3: Smoothing and gradient modification.}
To avoid abrupt changes in the boost signal, we apply temporal smoothing to the scales. We use a separate exponential moving average (EMA), distinct from the probe-accuracy EMA in Eq.~\eqref{eq:ema_probe}, though our analysis (Propositions~\ref{prop:attenuation} and~\ref{prop:convergence}) is stated specifically for this EMA:
\begin{equation}
\label{eq:ema_scale}
\bar{s}_m \leftarrow \mu \cdot s_m + (1 - \mu) \cdot \bar{s}_m,
\end{equation}
where $\mu \in (0,1)$ is the EMA smoothing coefficient on the boost scales. The smoothed scales are applied to the encoder gradients \emph{after} the backward pass and \emph{before} the optimizer step:
\begin{equation}
\label{eq:gradient_mod}
\frac{\partial \loss}{\partial \theta_m} \leftarrow \bar{s}_m \cdot \frac{\partial \loss}{\partial \theta_m}.
\end{equation}
%

\subsection{Composability with gradient throttling}
\label{sec:compose}

Because PGGB amplifies weaker modalities without modifying the dominant modality's gradient, it is mechanistically compatible with synchronous balancing methods such as gradient throttling, loss reweighting, and Pareto rescaling, with the empirical magnitude of the compound gain depending on the specific base method (Section~\ref{sec:main_results}, Appendix~\ref{app:composability}). The main experiments compose with OGM-GE~\citep{peng2022ogmge}, and Appendix~\ref{app:composability} validates composition with four additional methods.

OGM-GE damps the dominant modality's gradient by a \emph{coupled} signal (inter-modal gradient-magnitude ratio $\rho$ derived from $\loss_{\text{fusion}}$), whereas our boost uses a \emph{decoupled} signal (probe accuracy on stop-gradient features $\sg[\mathbf{z}_m]$). The two mechanisms compose multiplicatively on encoder gradients:
\begin{equation}
\label{eq:combined}
\frac{\partial \loss}{\partial \theta_m} \leftarrow \bar{s}_m \cdot \omega_m^{\text{ogm}} \cdot \frac{\partial \loss}{\partial \theta_m},
\end{equation}
where $\omega_m^{\text{ogm}} \leq 1$ is the OGM-GE throttle and $\bar{s}_m \geq 1$ is our boost scale. 

Following established practice for auxiliary unimodal supervision~\citep{guo2024cggm}, we optionally include unimodal auxiliary losses, yielding the total objective
\begin{equation}
\label{eq:total_loss}
\loss = \loss_{\text{fusion}} + \gamma \sum_{m=1}^{M} \ell_{\text{aux}}\!\bigl(g_m(\mathbf{z}_m),\; y\bigr),
\end{equation}
where $\gamma$ is a scalar weight on the per-modality auxiliary terms (set in Section~\ref{sec:experiments}). $\ell_{\text{aux}}$ matches the form of $\ell$ (Eq.~\eqref{eq:joint_loss}) for each benchmark.

\subsection{Safety properties}
\label{sec:convergence}

We establish three safety properties: bounded scaling, self-attenuation, and a standard-SGD descent bound. These are structural guarantees that the method does not diverge or harm low-imbalance training, not mechanism-specific convergence results. A rate-level analysis tying gap closure to probe dynamics is left for future work. Full proofs are in Appendix~\ref{app:proofs}.

\begin{assumption}[Smoothness]
\label{asm:smooth}
The loss $\loss(\theta)$ is $L$-smooth: $\|\nabla \loss(\theta) - \nabla \loss(\theta')\| \leq L\|\theta - \theta'\|$ for all $\theta, \theta'$.
\end{assumption}

\begin{assumption}[Bounded variance]
\label{asm:variance}
The stochastic gradient $\mathbf{g}(\theta)$ satisfies $\mathbb{E}[\mathbf{g}(\theta)] = \nabla \loss(\theta)$ and $\mathbb{E}[\|\mathbf{g}(\theta) - \nabla \loss(\theta)\|^2] \leq \sigma^2$ for all $\theta$.
\end{assumption}

\begin{proposition}[Bounded scaling]
\label{prop:bounded}
For all modalities $m$ and all iterations $t$, the smoothed boost scale satisfies $\bar{s}_m^{(t)} \in [1, s_{\max}]$.
\end{proposition}

\begin{proposition}[Self-attenuation]
\label{prop:attenuation}
For any $\Delta \geq 0$, if the smoothed probe accuracies are $\Delta$-balanced ($\max_j \bar{P}_j - \min_j \bar{P}_j \leq \Delta$) at iteration $t$, then each modality's instantaneous boost scale satisfies $|s_m^{(t)} - 1| \leq \alpha \Delta / (\Delta + \epsilon)$, and the EMA-smoothed scale converges geometrically to unity at rate $(1-\mu)$ as long as this condition persists.
\end{proposition}

\begin{proposition}[Descent bound]
\label{prop:convergence}
Under Assumptions~\ref{asm:smooth} and~\ref{asm:variance}, SGD with PGGB and constant learning rate $\eta \leq 1/(L \cdot s_{\max}^2)$ satisfies
\begin{equation}
\label{eq:convergence}
\frac{1}{T}\sum_{t=0}^{T-1} \mathbb{E}\bigl[\|\nabla \loss(\theta_t)\|^2\bigr]
\;\leq\; \frac{2\,\Delta_{\loss}}{\eta\, T} \;+\; L\,\eta\, s_{\max}^2\, \sigma^2,
\end{equation}
where $\Delta_{\loss} = \loss(\theta_0) - \loss^{*}$ is the initial optimality gap and $\loss^{*}$ is a lower bound on $\loss$.
\end{proposition}

The bound matches the standard nonconvex SGD rate~\citep{ghadimi2013sgd} up to an $s_{\max}^2$ factor in the variance term, a controlled increase ($s_{\max}=2$), and Proposition~\ref{prop:attenuation} ensures this cost is incurred only when imbalance is present. The smoothness constant $L$ in Assumption~\ref{asm:smooth} is loss-dependent, determined by the choice of $\ell$ in Eq.~\eqref{eq:joint_loss}; in our experiments, the per-task learning rate (Section~\ref{sec:experiments}) is selected consistent with the descent inequality for the corresponding $\ell$.

\paragraph{Scope in the high-imbalance regime.}
Proposition~\ref{prop:convergence} is uniform in the imbalance level: it holds for any utilization gap, provided the step size satisfies $\eta \le 1/(L s_{\max}^2)$. What limits its reach is the optimizer rather than the imbalance, since the proof assumes SGD. On high-imbalance benchmarks trained with SGD, the guarantee is therefore that PGGB preserves standard descent with gradient variance inflated by at most $s_{\max}^2 = 4$. The sentiment and text-image benchmarks are trained with Adam (Section~\ref{sec:experiments}) and fall outside this assumption, so our claims there rest on the measured results alone. Proposition~\ref{prop:attenuation} is the complementary guarantee at the balanced end, bounding the intervention by $|s_m - 1| \le \alpha \Delta / (\Delta + \epsilon)$, which tightens as the gap shrinks and gives $s_m = 1$ exactly at exact balance. For diagnosis, the operating-conditions analysis (Section~\ref{sec:operating_conditions}) identifies the regime in which composition becomes harmful, a thin data budget together with weak inherent asymmetry, and localizes the failure to the throttling component rather than to the boost. The monitored quantities $\delta$ and $\bar{s}_m$ can be logged during training as an early caution, since a small dominance gap signals that composing with throttling is not warranted, and an automated criterion for this decision remains future work.

\section{Experiments}
\label{sec:experiments}
\vspace{-1.2em}

\subsection{Experimental setup}
\label{sec:datasets}

\textbf{Datasets.}
We evaluate PGGB on eight benchmarks spanning three tasks and 2--4 modalities, reported as mean $\pm$ std over five seeds. Audio-visual and text-image benchmarks use two modalities, sentiment three (text, audio, vision), and BraTS~2021 four MRI sequences. \emph{Audio-visual:} CREMA-D~\citep{cao2014cremad} (3-frame at 3 fps), AVE~\citep{tian2018ave}, Kinetics-Sounds (KS)~\citep{arandjelovic2017kinetics}. \emph{Sentiment:} CH-SIMS~\citep{yu2020chsims}, CMU-MOSI~\citep{zadeh2016mosi}. \emph{Text-image:} Twitter15~\citep{yu2019tombert, zhang2018adaptive}, Sarcasm~\citep{cai2019sarcasm}. \emph{Segmentation:} BraTS~2021~\citep{baid2021brats}. Encoders follow task grouping (ResNet-18 for audio-visual, MLPs on pre-extracted features for sentiment and text-image, DeepLab v3+/ResNet-101 for BraTS), with splits and caveats in Appendix~\ref{app:datasets}. We assume each modality is task-relevant, consistent with these benchmarks' established multimodal use; PGGB does not validate relevance and would amplify a noise-only modality if one were present.

\textbf{Baselines.}
\label{sec:baselines}
We compare experiment results against eight methods: \textbf{Baseline} (joint training, no balancing), \textbf{OGM-GE}~\citep{peng2022ogmge} (discrepancy-ratio gradient throttling), \textbf{MMPareto}~\citep{wei2024mmpareto} (Pareto-optimal gradient weights), \textbf{AGM}~\citep{li2023agm} (Shapley-based adaptive gradient modulation), \textbf{G-Blend}~\citep{wang2020gblending} (OG-ratio loss weighting), \textbf{CGGM}~\citep{guo2024cggm} (classifier-guided magnitude/direction modulation\footnote{Adapted from its original Transformer setting to our CNN/MLP pipeline. The reproduction shows consistent regression across all eight benchmarks, including off-table Food101 ($-37$~pp vs.\ baseline, Appendix~\ref{app:datasets}), which we attribute to backbone mismatch rather than the underlying method.}), \textbf{InfoReg}~\citep{huang2025inforeg} (unimodal information regularization), and \textbf{MILES}~\citep{guerra2025miles} (per-modality learning-rate adjustment). InfoReg and MILES are evaluated on CREMA-D only, as their official implementations target audio-visual classification. MMPareto, AGM, and G-Blend are evaluated on all classification benchmarks but not on BraTS~2021, as their official implementations target classification losses and do not support dense-prediction multi-class Dice objectives out of the box.

\textbf{Reproduction protocol.} All baselines are re-run under a matched pipeline (shared encoders, 3-frame CREMA-D, identical augmentation and optimizer) for fair comparison. OGM-GE ($69.14\%$) is $3$--$6$~pp below the $72$--$75\%$ in follow-up work using different frame sampling, AGM ($57.42\%$) uses our ResNet-18 rather than its original backbone, MILES is ported to our matched ResNet-18 pipeline from its specialized backbone, and CGGM from its Transformer setting. Sub-baseline AGM/MILES numbers plausibly reflect 3-frame protocol sensitivity, and published configurations are retained without retuning.

\textbf{Implementation details.}
Classification uses concatenation fusion, batch size 64, 100 epochs. Audio-visual datasets use SGD (learning rate $0.001$, weight decay $10^{-4}$) with StepLR, sentiment and text-image use Adam (learning rate $0.001$) with StepLR (step $40$), and BraTS uses SGD with cosine scheduling (base learning rate $0.01$, 5-epoch warmup) and batch size $12$. PGGB uses identical hyperparameters across all datasets: $K{=}20$, $\beta{=}0.1$, $\mu{=}0.3$, $s_{\max}{=}2.0$, $\gamma{=}1.0$, and $\alpha \in \{0.5, 0.75\}$ for the standalone and OGM-GE-combined variants respectively. Probes are trained with Adam (learning rate $10^{-3}$).

\textbf{Computational overhead.} Each linear probe $\probe_m\colon \mathbb{R}^d \to \mathbb{R}^C$ adds at most a few thousand parameters per modality, negligible relative to the encoders (e.g., ResNet-18 at $\sim$11M parameters). Probe training every $K{=}20$ iterations on a half mini-batch and the multiplicative gradient scaling at the optimizer step together add approximately $1\%$ wall-clock overhead relative to the baseline. The EMA state $\{\bar{P}_m, \bar{s}_m\}$ requires $O(M)$ additional scalars.

\textbf{Imbalance categorization.}
\label{sec:imbalance_cat}
We categorize benchmarks by the utilization gap $\delta$ (Eq.~\eqref{eq:utilization_gap}) measured during baseline training, with per-dataset values in Appendix~\ref{app:delta}: \emph{high-imbalance} ($\delta > 0.15$, CREMA-D, CMU-MOSI, CH-SIMS) and \emph{low-imbalance} ($\delta \leq 0.15$, AVE, KS, Twitter15, Sarcasm). Here $\delta$ is the final-epoch EMA-smoothed probe-accuracy gap under baseline training with no intervention, averaged over five seeds, rather than a statistic over the full training trajectory, and it is measured on the seven classification benchmarks. The threshold is not tuned. Sorting the measured values gives AVE $0.017$, KS $0.038$, Sarcasm $0.076$, Twitter15 $0.150$, CH-SIMS $0.178$, CMU-MOSI $0.222$, and CREMA-D $0.268$, with no benchmark falling between $0.150$ and $0.178$, so every threshold in that interval produces the identical categorization and no result depends on the specific value $0.15$. Because $\delta$ is measured on baseline training alone, before any method is applied, the categorization is fixed independently of any method's results. Twitter15 ($0.150 \pm 0.096$) and CH-SIMS ($0.178 \pm 0.086$) lie closest to the threshold and carry the largest seed variances, so we flag both as boundary cases. BraTS~2021 is reported separately as a dense-prediction segmentation benchmark. The high-imbalance group is heterogeneous in source: CREMA-D imbalance arises from audio-dominance in emotion recognition~\citep{peng2022ogmge}, whereas CMU-MOSI/CH-SIMS imbalance reflects text-sufficiency, with text-only baselines within roughly $1$~pp of multimodal accuracy on CH-SIMS~\citep{liu2022chsimsv2} and only a $+2.7$~pp multimodal gain on CMU-MOSI~\citep{wei2025dgl}. On these text-dominant benchmarks, balancing methods report only modest absolute gains~\citep{guo2024cggm,wei2024opm}, motivating within-seed-std competitiveness as the relevant metric.

\begin{table}[H]
\centering
\caption{Main results across eight benchmarks. We report accuracy (\%) for classification datasets and average Dice (\%) for BraTS~2021. All values are mean $\pm$ std over 5 seeds. \textbf{Bold} indicates the best result per dataset. Imbalance categorization is described in Section~\ref{sec:imbalance_cat}. Baseline denotes joint training with unmodified gradients, not a published OGM-GE result. All methods are re-run under the matched pipeline of Section~\ref{sec:datasets}, in which CREMA-D uses 3 frames. OGM-GE's published 1-frame CREMA-D result ($61.9\%$)~\citep{peng2022ogmge} is reproduced at $62.47 \pm 1.42\%$ in Table~\ref{tab:ablation}.}
\label{tab:main_results}
\vspace{0.5em}
\resizebox{\textwidth}{!}{%
\begin{tabular}{l cccccccc}
\toprule
\textbf{Method} & CREMA-D & AVE & KS & CMU-MOSI & Twitter15 & Sarcasm & CH-SIMS & BraTS~2021 \\
\midrule
Baseline             & 61.59{\scriptsize$\pm$0.80} & 86.54{\scriptsize$\pm$0.42} & 79.05{\scriptsize$\pm$0.40} & 72.42{\scriptsize$\pm$0.48} & 66.60{\scriptsize$\pm$0.57} & 82.40{\scriptsize$\pm$0.17} & 70.42{\scriptsize$\pm$0.29} & 85.77{\scriptsize$\pm$0.62} \\
OGM-GE               & 69.14{\scriptsize$\pm$1.13} & 86.96{\scriptsize$\pm$0.71} & 77.25{\scriptsize$\pm$0.79} & \textbf{72.68}{\scriptsize$\pm$0.89} & 66.33{\scriptsize$\pm$0.21} & 81.81{\scriptsize$\pm$0.23} & \textbf{72.47}{\scriptsize$\pm$0.70} & 86.21{\scriptsize$\pm$0.17} \\
MMPareto              & 65.51{\scriptsize$\pm$0.87} & 86.37{\scriptsize$\pm$0.33} & 78.21{\scriptsize$\pm$0.38} & \textbf{72.68}{\scriptsize$\pm$0.79} & 66.81{\scriptsize$\pm$0.47} & 82.27{\scriptsize$\pm$0.12} & 70.20{\scriptsize$\pm$0.71} & --- \\
AGM                   & 57.42{\scriptsize$\pm$0.73} & 84.42{\scriptsize$\pm$0.26} & 77.85{\scriptsize$\pm$0.30} & 72.50{\scriptsize$\pm$0.62} & 66.79{\scriptsize$\pm$0.47} & 82.32{\scriptsize$\pm$0.16} & 69.28{\scriptsize$\pm$0.67} & --- \\
G-Blend               & 61.10{\scriptsize$\pm$1.87} & 87.09{\scriptsize$\pm$0.29} & 77.75{\scriptsize$\pm$1.13} & 72.36{\scriptsize$\pm$0.66} & 66.66{\scriptsize$\pm$0.57} & 82.35{\scriptsize$\pm$0.26} & 70.15{\scriptsize$\pm$0.66} & --- \\
CGGM                  & 50.22{\scriptsize$\pm$1.39} & 76.72{\scriptsize$\pm$0.42} & 73.18{\scriptsize$\pm$0.27} & 59.45{\scriptsize$\pm$0.36} & 62.41{\scriptsize$\pm$0.36} & 80.71{\scriptsize$\pm$0.27} & 68.05{\scriptsize$\pm$0.46} & 81.13{\scriptsize$\pm$1.34} \\
InfoReg               & 67.72{\scriptsize$\pm$0.83} & --- & --- & --- & --- & --- & --- & --- \\
\midrule
PGGB ($\alpha{=}0.5$)     & 62.72{\scriptsize$\pm$1.65} & \textbf{87.41}{\scriptsize$\pm$0.26} & \textbf{79.17}{\scriptsize$\pm$0.97} & 71.89{\scriptsize$\pm$0.82} & \textbf{66.85}{\scriptsize$\pm$0.50} & \textbf{82.44}{\scriptsize$\pm$0.35} & 69.80{\scriptsize$\pm$0.80} & 85.98{\scriptsize$\pm$1.15} \\
PGGB+OGM-GE ($\alpha{=}0.75$) & \textbf{71.45}{\scriptsize$\pm$1.71} & 87.23{\scriptsize$\pm$0.58} & 77.33{\scriptsize$\pm$0.64} & 72.60{\scriptsize$\pm$0.94} & 66.59{\scriptsize$\pm$0.51} & 81.76{\scriptsize$\pm$0.08} & 72.43{\scriptsize$\pm$0.65} & \textbf{86.49}{\scriptsize$\pm$0.21} \\
\bottomrule
\end{tabular}%
}
\end{table}

\subsection{Main results}
\label{sec:main_results}
\vspace{-1em}

Table~\ref{tab:main_results} reports results across eight benchmarks. On high-imbalance CREMA-D, PGGB+OGM-GE attains $71.45\%$, $+2.31$~pp over OGM-GE and $+9.86$~pp over baseline, the best among competitors. A three-way decomposition in \S\ref{sec:ablations} isolates this gain to the boost actuation, with the decoupled monitor causally inert at $\alpha{=}0$. PGGB alone ($+1.13$~pp) is insufficient under severe imbalance, confirming that two-sided intervention is required. On the four low-imbalance benchmarks (AVE, KS, Twitter15, Sarcasm, with cross-class MSLR comparison in Appendix~\ref{app:mslr}), PGGB alone is the best-performing method on all four, whereas OGM-GE often regresses (KS $-1.80$~pp, Twitter15 $-0.27$~pp, Sarcasm $-0.59$~pp), consistent with self-attenuation as $\delta \to 0$ (\S\ref{sec:boost}). On the high-imbalance text-dominant sentiment benchmarks (CMU-MOSI, CH-SIMS), where text alone is near-multimodally sufficient and prior balancing methods report only modest absolute gains~\citep{liu2022chsimsv2,wei2025dgl}, PGGB+OGM-GE remains within seed standard deviation of the strongest baseline (MOSI $-0.08$~pp, CH-SIMS $-0.04$~pp). On BraTS~2021 (DeepLab v3+/ResNet-101, Dice loss), PGGB+OGM-GE yields $86.49\%$ Dice ($+0.72$~pp), extending the mechanism to dense prediction. The composability claim further extends on AVE to two of four non-OGM-GE methods tested (Appendix~\ref{app:composability}). A comparison against OPM~\citep{wei2024opm}, whose single-layer-fusion dependency requires a separate protocol, is in Appendix~\ref{app:opm_comparison}.
\vspace{-1.2em}

\subsection{Ablation studies}
\label{sec:ablations}

We ablate on CREMA-D with single-frame visual input, which widens the modality gap and amplifies method differences (Table~\ref{tab:ablation}).

\begin{table}[H]
\centering
\caption{Ablation on CREMA-D (1-frame). Accuracy (\%), mean $\pm$ std over 5 seeds. The upper row reports the baseline. Lower rows isolate single-mechanism interventions (boost only, OGM-GE only) and their combinations.}
\label{tab:ablation}
\vspace{0.5em}
\begin{tabular}{lc}
\toprule
\textbf{Method} & \textbf{Accuracy (\%)} \\
\midrule
Baseline                              & 60.30{\scriptsize$\pm$0.70} \\
\midrule
PGGB ($\alpha{=}0.5$)                & 60.46{\scriptsize$\pm$0.85} \\
OGM-GE only                          & 62.47{\scriptsize$\pm$1.42} \\
PGGB+OGM-GE ($\alpha{=}0.5$)         & 62.37{\scriptsize$\pm$0.57} \\
PGGB+OGM-GE ($\alpha{=}0.75$)        & \textbf{62.69}{\scriptsize$\pm$0.22} \\
\bottomrule
\end{tabular}
\end{table}

Among single-mechanism interventions, OGM-GE alone yields $+2.17$~pp versus PGGB alone $+0.16$~pp, consistent with prior findings that throttling is the more impactful individual intervention~\citep{wei2024opm}. Combining PGGB with OGM-GE at $\alpha{=}0.75$ achieves the best result ($62.69\%$) with a sixfold reduction in std ($0.22\%$ vs.\ $1.42\%$), indicating that decoupled probe monitoring provides a stabilizing signal complementing the noisier gradient-ratio throttling.
\vspace{-1em}

\paragraph{Isolating the monitor from the boost.} We run three 5-seed decompositions on CREMA-D 3-frame (Table~\ref{tab:main_results} protocol): (i) $\alpha{=}0$ with probes active ($69.14 \pm 1.13\%$, reported as OGM-GE alone in Table~\ref{tab:main_results}), (ii) an independent 5-seed replication of the same $\alpha{=}0$ condition ($69.14 \pm 1.18\%$), and (iii) $\alpha{=}0.75$ ($71.45 \pm 1.71\%$). Conditions (i) and (ii) are operationally identical (identical $\alpha$, identical probe pipeline) and yield the same mean within $0.00$~pp, confirming estimate stability. The $+2.31$~pp lift at (iii) isolates boost actuation: it cannot be attributed to probe-pipeline side-effects matched across conditions (probe training on stop-gradient features, optimizer steps on probe parameters, and EMA state updates). Appendix~\ref{app:perseed} extends this contrast to 15 seeds per arm, giving $+1.60$~pp with a confidence interval excluding zero. Only the multiplicative scale $s_m$ differs between $\alpha{=}0$ and $\alpha{=}0.75$.
\vspace{-1em}

\subsubsection{Operating conditions ablation}
\label{sec:operating_conditions}
\vspace{-1em}

\begin{figure}[H]
\centering
\begin{subfigure}[b]{0.55\textwidth}
\centering
\includegraphics[width=\linewidth]{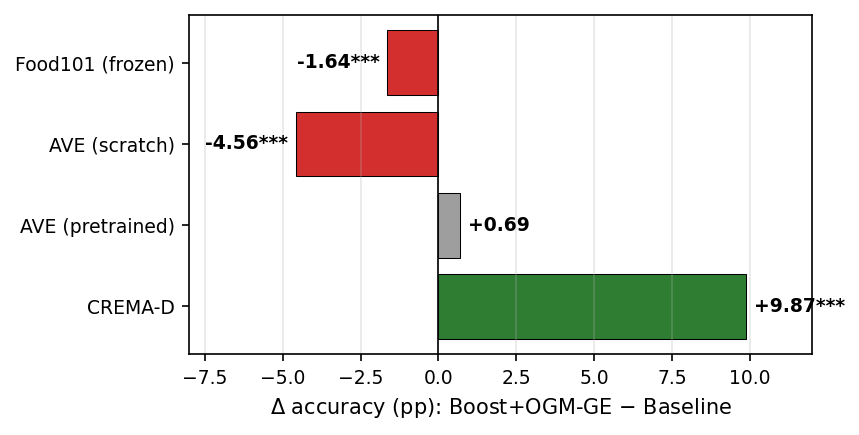}
\caption{$\Delta$ accuracy (PGGB+OGM-GE vs.\ baseline) across four regimes.}
\label{fig:operating_conditions}
\end{subfigure}
\hfill
\begin{subfigure}[b]{0.43\textwidth}
\centering
\includegraphics[width=\linewidth]{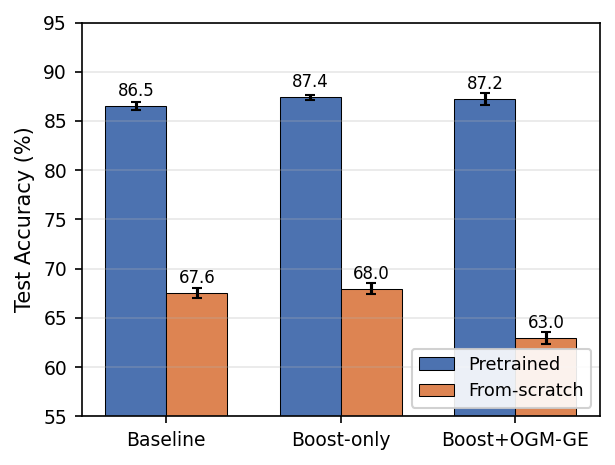}
\caption{AVE test accuracy: pretrained vs.\ from-scratch encoders.}
\label{fig:ave_contrast}
\end{subfigure}
\caption{Operating-conditions ablation (5 seeds). (a) PGGB alone self-attenuates toward baseline when Table~\ref{tab:operating_conditions} conditions are violated, whereas PGGB+OGM-GE inherits OGM-GE's regime-dependent failure modes (AVE from-scratch is the clearest case). (b) With data and architecture fixed, flipping initialization from pretrained to random shifts PGGB+OGM-GE from $+0.69$ to $-4.56$~pp ($p{=}0.0001$) while PGGB alone stays positive, localizing sensitivity to the OGM-GE component.}
\label{fig:operating_conditions_combined}
\end{figure}

A complementary question to \emph{which} component drives the CREMA-D gain is \emph{when} the method is effective. We characterize four regimes along three axes (feature learning, encoder initialization, data-to-architecture ratio) in Table~\ref{tab:operating_conditions} and Figure~\ref{fig:operating_conditions_combined}.
\vspace{-1.2em}

\begin{table}[H]
\centering
\caption{Operating conditions across four regimes. Columns: trainable encoders, modality asymmetry (pretraining or LR differences), adequate data-to-architecture ratio, and complementary modalities. $\Delta$: accuracy change (pp), PGGB+OGM-GE vs.\ baseline. Significance: \textsuperscript{***}$p<0.001$ (Welch's t-test, 5 seeds per condition). AVE (from-scratch) is the clearest failure case ($-4.56$ pp), with the pretrained$\leftrightarrow$scratch contrast localizing sensitivity to the OGM-GE component. Per-regime analysis in Appendix~\ref{app:opcond}.}
\label{tab:operating_conditions}
\vspace{0.3em}
\footnotesize
\setlength{\tabcolsep}{4pt}
\begin{tabular}{lccccc}
\toprule
\textbf{Regime} & \textbf{Trainable} & \textbf{Asymmetry} & \textbf{Adequate data} & \textbf{Compl.} & $\boldsymbol{\Delta}$ \\
\midrule
CREMA-D             & $\checkmark$ & $\checkmark$            & $\checkmark$         & $\checkmark$ & $+9.86$\textsuperscript{***} \\
AVE (pretrained)    & $\checkmark$ & $\checkmark$ (pretrain) & $\checkmark$         & $\checkmark$ & $+0.69$ \\
AVE (from-scratch)  & $\checkmark$ & $\times$                & $\times$ ($\sim 4$K) & $\checkmark$ & $-4.56$\textsuperscript{***} \\
Food101 (frozen)    & $\times$     & ---                     & $\checkmark$         & $\checkmark$ & $-1.64$ \\
\bottomrule
\end{tabular}
\end{table}
\vspace{-1.8em}

\vspace{-1.2em}

\paragraph{Encoder initialization and scope.}
Fixing dataset and architecture but varying encoder initialization gives the sharpest contrast (Figure~\ref{fig:ave_contrast}). PGGB alone stays positive in both regimes ($+0.87$/$+0.42$~pp, both $p<0.05$), while PGGB+OGM-GE flips sign ($+0.69 \to -4.56$~pp, $p{=}0.0001$). ImageNet pretraining installs an asymmetry aligned with OGM-GE's throttling assumption. Under from-scratch initialization both encoders are simultaneously under-trained on $\sim$$4$K samples, so suppressing either destabilizes the joint dynamic. This localizes the failure to the OGM-GE component (Appendix~\ref{app:opcond}).

\subsection{Analysis}
\label{sec:analysis}
\vspace{-1.2em}

\begin{figure}[H]
\centering
\includegraphics[width=1.0\textwidth]{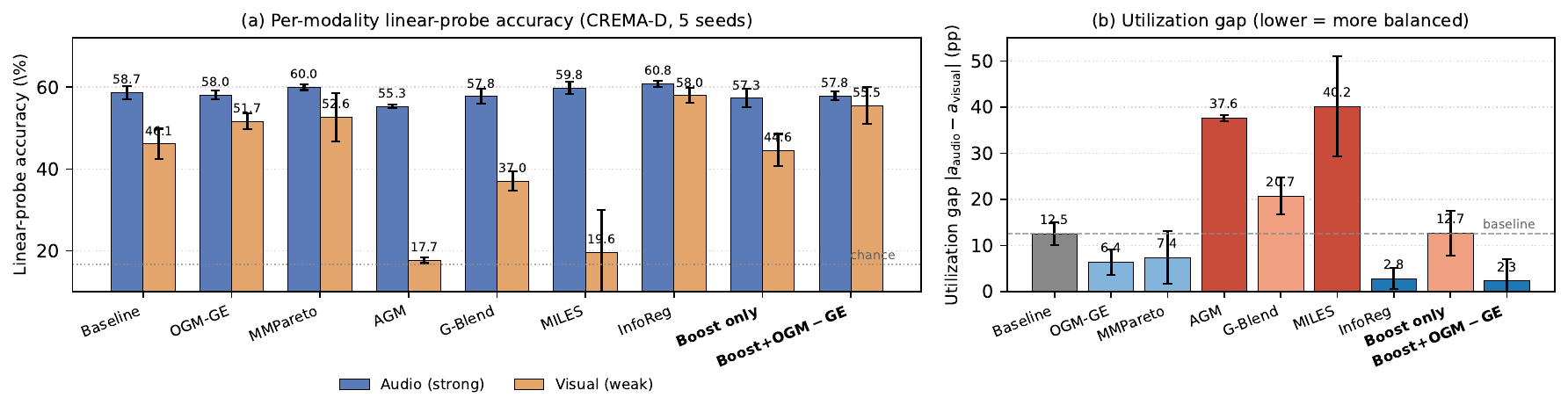}
\caption{Per-modality diagnostics on CREMA-D (5 seeds). \textbf{(a)} Post-hoc linear-probe accuracy $P_m^{\text{post}}$ (Appendix~\ref{app:posthoc_probe_table}). Dotted line marks chance ($1/6$). AGM and MILES collapse the weak probe. \textbf{(b)} Utilization gap $|P_a^{\text{post}} - P_v^{\text{post}}|$. PGGB+OGM-GE attains $2.31$~pp ($5.4\times$ reduction vs.\ baseline gap). Error bars: $\pm 1$ std.}
\label{fig:probe_accuracy}
\end{figure}
\vspace{-1.2em}

We probe per-modality utilization with linear classifiers trained on frozen encoder features. Post-hoc linear probes (Appendix~\ref{app:posthoc_probe_table}, Figure~\ref{fig:probe_accuracy}b) show PGGB+OGM-GE attains the smallest utilization gap ($2.31$~pp): the weak probe rises $+9.41$~pp, the strong falls only $-0.81$~pp, while AGM, MILES, and G-Blend degrade the weak probe. Extension to $M$ modalities uses pairwise weakness (Eq.~\eqref{eq:weakness}).

\FloatBarrier
\enlargethispage{2\baselineskip}
\section{Conclusion}
\label{sec:conclusion}

We propose PGGB, a method for balanced multimodal learning that amplifies the contribution of weaker modalities while preserving stronger ones, composing naturally with existing throttling techniques. PGGB alone is the best-performing method on all four low-imbalance benchmarks. On high-imbalance CREMA-D, composition with OGM-GE yields $+2.31$~pp over the strongest baseline, and on the text-dominant sentiment benchmarks (CMU-MOSI, CH-SIMS) the same composition remains within seed standard deviation of the strongest baseline. On medical segmentation (BraTS~2021), it matches the strongest available baseline ($+0.28$~pp, within seed variance).

\emph{Limitations.} Per-setting tuning is required, probes add small compute overhead, and composition with throttling can regress when the dominance gap is small (worst case AVE-from-scratch $-4.56$~pp, \S\ref{sec:operating_conditions}), a composition our propositions do not cover. Those propositions are safety properties: they bound the intervention and, under Proposition~\ref{prop:convergence}'s conditions, preserve standard descent at any imbalance level, but do not guarantee gap closure, which is established only empirically. The two benchmarks nearest the threshold, Twitter15 and CH-SIMS, carry the largest seed variances, so their categorization is provisional (Appendix~\ref{app:delta_stability}). Finally, PGGB does not distinguish undertrained modalities from intrinsically less task-informative ones: composition recovers the former (CREMA-D), whereas on the sentiment benchmarks the weak-modality representations are essentially unchanged (Appendix~\ref{app:posthoc_sentiment}), a diagnosis available only post hoc. Actuation is also optimizer-dependent: scaling reaches the updates under SGD, whereas Adam's second-moment normalization absorbs most of it, so measured gains concentrate on the SGD pipelines (Appendix~\ref{app:optimizer}), leaving optimizer-state-aware actuation to future work.

\emph{Future work.} Probe-driven adaptive composition that engages throttling only when the dominance gap warrants, extension beyond classification and segmentation. The operating conditions of \S\ref{sec:operating_conditions} should precede deployment.

\begin{ack}
This work received no third-party funding. The authors
declare no competing interests. No author holds a financial relationship of any
kind that could be perceived to influence the work
reported in this paper.
\end{ack}

\newpage
\bibliographystyle{plainnat}
\bibliography{references}


\clearpage
\appendix

\section{Proofs of convergence properties}
\label{app:proofs}

We restate and prove each proposition from Section~\ref{sec:convergence}. All notation follows the main text: $\loss$ denotes the training loss, $\mathbf{g}(\theta)$ a stochastic gradient, $\bar{s}_m$ the EMA-smoothed boost scale for modality $m$, and $s_{\max}$ the scale cap.

\begin{proposition}[Bounded scaling, restated]
For all modalities $m$ and all iterations $t \geq 0$, the smoothed boost scale satisfies $\bar{s}_m^{(t)} \in [1, s_{\max}]$.
\end{proposition}

\begin{proof}
We proceed by induction on $t$. At initialization, $\bar{s}_m^{(0)} = 1 \in [1, s_{\max}]$, establishing the base case.

For the inductive step, suppose $\bar{s}_m^{(t)} \in [1, s_{\max}]$. By Eq.~\eqref{eq:weakness}, the weakness score satisfies $w_m \in [0, 1]$ since the numerator $\max_j \bar{P}_j - \bar{P}_m$ lies in $[0, \max_j \bar{P}_j - \min_j \bar{P}_j]$ and the denominator equals $\max_j \bar{P}_j - \min_j \bar{P}_j + \epsilon$. Substituting into Eq.~\eqref{eq:boost_scale}:
\begin{equation*}
s_m = \min(1 + \alpha \cdot w_m,\; s_{\max}) \in [1, s_{\max}],
\end{equation*}
since $w_m \geq 0$ implies $1 + \alpha \cdot w_m \geq 1$, and the $\min$ operation enforces the upper bound.

The EMA update (Eq.~\eqref{eq:ema_scale}) gives:
\begin{equation*}
\bar{s}_m^{(t+1)} = \mu \cdot s_m + (1 - \mu) \cdot \bar{s}_m^{(t)}.
\end{equation*}
Since $\mu \in (0, 1)$, this is a convex combination of $s_m \in [1, s_{\max}]$ and $\bar{s}_m^{(t)} \in [1, s_{\max}]$ (by the inductive hypothesis), so $\bar{s}_m^{(t+1)} \in [1, s_{\max}]$.
\end{proof}

\begin{proposition}[Self-attenuation, restated]
For any $\Delta \geq 0$, if $\max_j \bar{P}_j^{(t)} - \min_j \bar{P}_j^{(t)} \leq \Delta$ at all iterations $t \geq t_0$, then
\begin{enumerate}
    \item $|s_m^{(t)} - 1| \leq \alpha \cdot \Delta / (\Delta + \epsilon)$ for all $m$ and all $t \geq t_0$, and
    \item $|\bar{s}_m^{(t)} - 1| \leq (1 - \mu)^{t - t_0} \cdot (s_{\max} - 1) + \alpha \cdot \Delta / (\Delta + \epsilon)$.
\end{enumerate}
In particular, when $\Delta \to 0$ we recover $s_m = 1$ exactly, and the EMA-smoothed scale converges to unity at rate $(1-\mu)$.
\end{proposition}

\begin{proof}
Let $\delta^{(t)} := \max_j \bar{P}_j^{(t)} - \min_j \bar{P}_j^{(t)}$. By assumption $\delta^{(t)} \leq \Delta$ for all $t \geq t_0$. From Eq.~\eqref{eq:weakness}, the weakness score satisfies
\begin{equation*}
w_m = \frac{\max_j \bar{P}_j - \bar{P}_m}{\max_j \bar{P}_j - \min_j \bar{P}_j + \epsilon} = \frac{\max_j \bar{P}_j - \bar{P}_m}{\delta + \epsilon} \leq \frac{\Delta}{\Delta + \epsilon},
\end{equation*}
since the numerator is bounded by $\delta \leq \Delta$ and the denominator is non-decreasing in $\delta$. Then by Eq.~\eqref{eq:boost_scale},
\begin{equation*}
s_m = \min(1 + \alpha w_m,\, s_{\max}) \in \bigl[1,\, 1 + \alpha \cdot \tfrac{\Delta}{\Delta + \epsilon}\bigr],
\end{equation*}
which gives part (1) immediately. For part (2), the EMA update (Eq.~\eqref{eq:ema_scale}) is
\begin{equation*}
\bar{s}_m^{(t+1)} - 1 = \mu (s_m^{(t)} - 1) + (1 - \mu)(\bar{s}_m^{(t)} - 1),
\end{equation*}
so by the triangle inequality and induction from $t_0$,
\begin{equation*}
|\bar{s}_m^{(t)} - 1| \leq (1-\mu)^{t - t_0} |\bar{s}_m^{(t_0)} - 1| + \mu \sum_{k=0}^{t - t_0 - 1} (1-\mu)^k \cdot \max_{t_0 \leq \tau < t} |s_m^{(\tau)} - 1|.
\end{equation*}
Using $|\bar{s}_m^{(t_0)} - 1| \leq s_{\max} - 1$ (Proposition~\ref{prop:bounded}), $|s_m^{(\tau)} - 1| \leq \alpha \Delta / (\Delta + \epsilon)$, and the partial sum $\mu \sum_{k=0}^{t-t_0-1} (1-\mu)^{k} = 1 - (1-\mu)^{t-t_0} \leq 1$, the bound follows. In the limit $\Delta \to 0$ the second term vanishes and the deviation decays geometrically at rate $(1-\mu)$.
\end{proof}

\begin{proposition}[Convergence, restated]
Under Assumptions~\ref{asm:smooth} and~\ref{asm:variance}, SGD with PGGB and constant learning rate $\eta \leq 1/(L \cdot s_{\max}^2)$ satisfies
\begin{equation*}
\frac{1}{T}\sum_{t=0}^{T-1} \mathbb{E}\bigl[\|\nabla \loss(\theta_t)\|^2\bigr]
\;\leq\; \frac{2\,\Delta_{\loss}}{\eta\, T} \;+\; L\,\eta\, s_{\max}^2\, \sigma^2,
\end{equation*}
where $\Delta_{\loss} = \loss(\theta_0) - \loss^{*}$.
\end{proposition}

\begin{proof}
Let $\tilde{\mathbf{g}}_t$ denote the boosted stochastic gradient at iteration $t$. Concatenating over all modality-specific parameter blocks, $\tilde{\mathbf{g}}_t$ has block $m$ equal to $\bar{s}_m^{(t)} \cdot \mathbf{g}_m(\theta_t)$. The parameter update is $\theta_{t+1} = \theta_t - \eta\,\tilde{\mathbf{g}}_t$.

\paragraph{Step 1: Descent lemma.}
By $L$-smoothness (Assumption~\ref{asm:smooth}):
\begin{equation}
\label{eq:descent}
\loss(\theta_{t+1}) \leq \loss(\theta_t) - \eta \langle \nabla \loss(\theta_t),\, \tilde{\mathbf{g}}_t \rangle + \frac{L\eta^2}{2}\|\tilde{\mathbf{g}}_t\|^2.
\end{equation}

\paragraph{Step 2: Bounding the inner product.}
Taking expectations conditioned on $\theta_t$ and using $\mathbb{E}[\mathbf{g}_m(\theta_t)] = \nabla_{\theta_m} \loss(\theta_t)$:
\begin{align}
\mathbb{E}\bigl[\langle \nabla \loss(\theta_t),\, \tilde{\mathbf{g}}_t \rangle\bigr]
&= \sum_m \bar{s}_m^{(t)} \cdot \|\nabla_{\theta_m} \loss(\theta_t)\|^2 \nonumber \\
&\geq \sum_m \|\nabla_{\theta_m} \loss(\theta_t)\|^2 = \|\nabla \loss(\theta_t)\|^2,
\label{eq:inner_bound}
\end{align}
where the inequality follows from $\bar{s}_m^{(t)} \geq 1$ (Proposition~\ref{prop:bounded}). Let $\mathcal{F}_t$ denote the $\sigma$-algebra generated by $\{\theta_0, \dots, \theta_t\}$ together with the probe parameters and their EMA state up to step $t$. By construction the probe parameters $\psi_m$ are updated only on stop-gradient features $\sg[\mathbf{z}_m]$ (no gradient path into $\theta$), and the scale EMA $\bar{s}_m^{(t)}$ is computed from probe-accuracy EMA values $\bar{P}_m$ refreshed at iteration boundaries (every $K$ steps). Since the boost scale $\bar{s}_m^{(t)}$ applied at iteration $t$ is the value most recently refreshed at the previous probe event (iteration $t' < t$ with $t' \equiv 0 \pmod K$), it is $\mathcal{F}_{t'}$-measurable and hence independent of the gradient noise at iteration $t$. Therefore $\bar{s}_m^{(t)}$ is $\mathcal{F}_t$-measurable, and conditioning on $\mathcal{F}_t$, the stochastic gradient $\mathbf{g}(\theta_t)$ is a fresh independent sample with $\mathbb{E}[\mathbf{g}(\theta_t) \mid \mathcal{F}_t] = \nabla\loss(\theta_t)$ and variance bounded per Assumption~\ref{asm:variance}.

\paragraph{Step 3: Bounding the second moment.}
By Proposition~\ref{prop:bounded}, $\bar{s}_m^{(t)} \leq s_{\max}$. Writing the stochastic-gradient deviation over the block decomposition of parameters as $\mathbf{g}(\theta_t) - \nabla \loss(\theta_t) = \bigl(\mathbf{g}_1(\theta_t) - \nabla_{\theta_1}\loss(\theta_t),\,\dots,\,\mathbf{g}_M(\theta_t) - \nabla_{\theta_M}\loss(\theta_t)\bigr)$, we have $\mathbb{E}[\|\mathbf{g} - \nabla\loss\|^{2}] = \sum_m \mathbb{E}[\|\mathbf{g}_m - \nabla_{\theta_m}\loss\|^{2}] =: \sum_m \sigma_m^{2}$, so Assumption~\ref{asm:variance} is equivalent to $\sum_m \sigma_m^{2} \leq \sigma^{2}$. Using this block-additive decomposition:
\begin{align}
\mathbb{E}\bigl[\|\tilde{\mathbf{g}}_t\|^2\bigr]
&= \sum_m (\bar{s}_m^{(t)})^2 \cdot \mathbb{E}\bigl[\|\mathbf{g}_m(\theta_t)\|^2\bigr] \nonumber \\
&\leq s_{\max}^2 \sum_m \bigl(\sigma_m^2 + \|\nabla_{\theta_m} \loss(\theta_t)\|^2\bigr) \nonumber \\
&\leq s_{\max}^2 \bigl(\sigma^2 + \|\nabla \loss(\theta_t)\|^2\bigr),
\label{eq:second_moment}
\end{align}
where $\sigma^2 \geq \sum_m \sigma_m^2$ is the total variance bound across modality blocks.

\paragraph{Step 4: Combining and telescoping.}
Substituting Eqs.~\eqref{eq:inner_bound} and~\eqref{eq:second_moment} into Eq.~\eqref{eq:descent} and taking full expectations:
\begin{equation*}
\mathbb{E}\bigl[\loss(\theta_{t+1})\bigr] \leq \mathbb{E}\bigl[\loss(\theta_t)\bigr]
- \eta\,\mathbb{E}\bigl[\|\nabla \loss(\theta_t)\|^2\bigr]
+ \frac{L\eta^2 s_{\max}^2}{2}\bigl(\sigma^2 + \mathbb{E}\bigl[\|\nabla \loss(\theta_t)\|^2\bigr]\bigr).
\end{equation*}
Rearranging and using the learning rate condition $\eta \leq 1/(L \cdot s_{\max}^2)$: since $\eta \leq 1/(L \cdot s_{\max}^{2})$ implies $L\eta s_{\max}^{2} \leq 1$, hence $L\eta s_{\max}^{2}/2 \leq 1/2$, which ensures $1 - \tfrac{L\eta\, s_{\max}^2}{2} \geq \tfrac{1}{2}$:
\begin{equation*}
\frac{\eta}{2}\,\mathbb{E}\bigl[\|\nabla \loss(\theta_t)\|^2\bigr]
\leq \mathbb{E}\bigl[\loss(\theta_t)\bigr] - \mathbb{E}\bigl[\loss(\theta_{t+1})\bigr]
+ \frac{L\eta^2 s_{\max}^2}{2}\,\sigma^2.
\end{equation*}
Summing over $t = 0, \ldots, T-1$ and dividing by $T$:
\begin{equation*}
\frac{1}{T}\sum_{t=0}^{T-1} \mathbb{E}\bigl[\|\nabla \loss(\theta_t)\|^2\bigr]
\leq \frac{2(\loss(\theta_0) - \mathbb{E}[\loss(\theta_T)])}{\eta\, T}
+ L\,\eta\, s_{\max}^2\, \sigma^2
\leq \frac{2\,\Delta_{\loss}}{\eta\, T}
+ L\,\eta\, s_{\max}^2\, \sigma^2,
\end{equation*}
where $\Delta_{\loss} = \loss(\theta_0) - \loss^{*}$ and the last inequality uses $\mathbb{E}[\loss(\theta_T)] \geq \loss^{*}$.

\paragraph{Interpretation.} Setting $\eta = \Theta(1/(s_{\max}\sqrt{T}))$ yields the rate $\mathcal{O}(s_{\max} \cdot \sigma / \sqrt{T} + s_{\max}^2 \cdot \Delta_{\loss} / T)$. For the default $s_{\max} = 2.0$, the effective variance increases by a factor of at most 4 relative to unmodified SGD. Proposition~\ref{prop:attenuation} guarantees that as modalities become $\Delta$-balanced, the boost scales approach unity within $\alpha \Delta / (\Delta + \epsilon)$ and this overhead vanishes in the limit $\Delta \to 0$, recovering the standard SGD convergence rate~\citep{ghadimi2013sgd} up to a constant variance inflation of at most $s_{\max}^2$.
\end{proof}

\subsection{Training algorithm pseudocode}
\label{app:algorithm}

Algorithm~\ref{alg:pgb} below specifies the full training pass referenced in Section~\ref{sec:boost}, making explicit how probe evaluation, weakness scoring, scale smoothing, and gradient boosting interleave with the forward and backward passes. The optional OGM-GE throttle line corresponds to the composition described in Section~\ref{sec:compose}.

\begin{algorithm}[H]
\caption{Probe-Guided Gradient Boosting (full training pass over $\mathcal{D}$). Notation: $P_m$ per-batch probe accuracy with smoothed estimate $\bar{P}_m$; $\phi$ fusion-head parameters; $\sg[\cdot]$ stop-gradient operator (\S\ref{sec:probes}). Auxiliary-loss weight $\gamma$ and optional OGM-GE throttle $\omega_m^{\text{ogm}} \leq 1$ are introduced in \S\ref{sec:compose}.}
\label{alg:pgb}
\begin{algorithmic}[1]
\REQUIRE Training data $\mathcal{D}$, encoders $\{\encoder_m\}_{m=1}^M$, fusion head $g$, unimodal heads $\{g_m\}$, probes $\{\probe_m\}$
\REQUIRE Boost strength $\alpha$, cap $s_{\max}$, smoothing coefficients $\beta$, $\mu$, probe interval $K$
\REQUIRE Optional: OGM-GE throttle module
\STATE Initialize $\bar{P}_m \leftarrow 0$, $\bar{s}_m \leftarrow 1$ for all $m$
\FOR{each mini-batch $(\mathbf{x}, y) \in \mathcal{D}$}
    \STATE \textbf{Forward:} $\mathbf{z}_m \leftarrow \encoder_m(\mathbf{x}_m;\,\theta_m)$ for all $m$
    \STATE \textbf{Loss:} $\loss \leftarrow \ell(g([\mathbf{z}_1;\ldots;\mathbf{z}_M]), y) + \gamma \sum_m \ell_{\text{aux}}(g_m(\mathbf{z}_m), y)$
    \STATE \textbf{Backward:} compute $\partial\loss/\partial\theta_m$ for all $m$
    \IF{OGM-GE enabled}
        \STATE Apply OGM-GE throttle: $\partial\loss/\partial\theta_m \leftarrow \omega_m^{\text{ogm}} \cdot \partial\loss/\partial\theta_m$
    \ENDIF
    \IF{iteration $\bmod\; K = 0$}
        \STATE Split $(\mathbf{x}, y)$ into train half $(\mathbf{x}^{\text{tr}}, y^{\text{tr}})$ and eval half $(\mathbf{x}^{\text{ev}}, y^{\text{ev}})$
        \FOR{each modality $m$}
            \STATE Train $\probe_m$ on $(\sg[\mathbf{z}_m^{\text{tr}}],\; y^{\text{tr}})$ with separate optimizer
            \STATE Evaluate: $P_m \leftarrow \text{Acc}(\probe_m(\sg[\mathbf{z}_m^{\text{ev}}]),\; y^{\text{ev}})$
        \ENDFOR
        \STATE Update $\bar{P}_m \leftarrow \beta \cdot P_m + (1-\beta) \cdot \bar{P}_m$ for all $m$
        \STATE Compute $w_m$ via Eq.~\eqref{eq:weakness}, $s_m$ via Eq.~\eqref{eq:boost_scale} for all $m$
        \STATE Update $\bar{s}_m \leftarrow \mu \cdot s_m + (1-\mu) \cdot \bar{s}_m$ for all $m$
    \ENDIF
    \STATE \textbf{Boost:} $\partial\loss/\partial\theta_m \leftarrow \bar{s}_m \cdot \partial\loss/\partial\theta_m$ for all $m$
    \STATE \textbf{Optimizer step:} update $\theta_m$, $\phi$, and $\{g_m\}$ parameters
\ENDFOR
\end{algorithmic}
\end{algorithm}


\section{Additional Experiments}
\label{app:datasets}

\subsection{Datasets}
\label{app:datasets_list}

\noindent\textbf{CREMA-D:} CREMA-D~\citep{cao2014cremad} is an audio-visual dataset designed for speech emotion recognition. It contains 7{,}442 video clips of 2--3 seconds from 91 actors speaking short scripted sentences, divided into 6{,}698 training samples and 744 testing samples. The dataset covers six emotions: angry, happy, sad, neutral, disgust, and fear. We encode audio as log-mel spectrograms and visual as 3 RGB frames sampled at 3 fps, passing both through ResNet-18 backbones trained from scratch. Available at \url{https://github.com/CheyneyComputerScience/CREMA-D}.

\noindent\textbf{AVE:} AVE~\citep{tian2018ave} is an audio-visual event classification dataset with 28 event categories, containing 4{,}143 10-second video clips. We use the standard split of 3{,}328 training and 402 testing clips. Audio spectrograms and 3 RGB frames are processed by ImageNet-pretrained ResNet-18 backbones with concatenation fusion. Available at \url{https://sites.google.com/view/audiovisualresearch}.

\noindent\textbf{Kinetics-Sounds (KS):} KS~\citep{arandjelovic2017kinetics} is a curated 31-class subset of the Kinetics action-recognition dataset whose actions are both visually and audibly recognizable. It contains roughly 25{,}000 10-second YouTube clips, split into 22{,}408 training and 3{,}000 testing clips. Encoder setup follows AVE (ImageNet-pretrained ResNet-18 backbones on audio spectrograms and 3 RGB frames). Available at \url{https://github.com/cvdfoundation/kinetics-dataset}.

\noindent\textbf{CH-SIMS:} CH-SIMS~\citep{yu2020chsims} is a Mandarin Chinese language-audio-visual sentiment dataset of 2{,}281 video segments. It is divided into 1{,}368 training, 456 validation, and 457 testing samples. We use pre-extracted Chinese BERT-base ($39 \times 768$) text, LibROSA ($400 \times 33$) audio, and OpenFace 2.0 ($55 \times 709$) visual features with the supplied 3-class sentiment labels (negative, neutral, positive). Features are z-score normalized per dimension and mean-pooled over the temporal axis before encoding. Each modality is processed by a two-layer MLP with 512-d hidden layer and dropout $0.3$. Available through the MMSA (MultiModal Sentiment Analysis) toolkit at \url{https://github.com/thuiar/MMSA}.

\noindent\textbf{CMU-MOSI:} CMU-MOSI~\citep{zadeh2016mosi} is an English language-audio-visual sentiment dataset with the same three modalities as CH-SIMS. It is divided into 1{,}283 training, 214 validation, and 686 testing samples. We use pre-extracted GloVe (300-d) text, COVAREP (74-d) audio, and FACET (35-d) visual features, each as 50-step sequences, with continuous sentiment scores which we binarize at zero into 2-class labels (negative, positive). Features are mean-pooled over the temporal axis before encoding. Encoder architecture is identical to CH-SIMS. Available at \url{https://github.com/CMU-MultiComp-Lab/CMU-MultimodalSDK}.

\noindent\textbf{Twitter15:} Twitter15~\citep{yu2019tombert, zhang2018adaptive} is a text-image sentiment dataset of tweets with associated images, covering three classes (positive, neutral, negative). It is divided into 3{,}179 training pairs, 1{,}122 validation pairs, and 1{,}037 testing pairs. We use MLP heads on frozen BERT-base (768-d) text and ImageNet ResNet-18 (512-d) image features, so only the heads are trainable and gradient-balancing methods act on them. Available at \url{https://github.com/jefferyYu/TomBERT}.

\noindent\textbf{Sarcasm:} The MMSD-v1 sarcasm dataset~\citep{cai2019sarcasm} is a text-image binary sarcasm detection benchmark of 24{,}635 image-text pairs, divided into 19{,}816 training, 2{,}410 validation, and 2{,}409 testing pairs. Encoder setup is identical to Twitter15 (MLP heads on frozen BERT and ResNet-18 features). Available at \url{https://github.com/headacheboy/data-of-multimodal-sarcasm-detection}.

\noindent\textbf{BraTS 2021:} BraTS~2021~\citep{baid2021brats} is a multimodal brain tumor segmentation dataset with 4 MRI modalities (FLAIR, T1ce, T1, T2) and 4 voxel-level class labels (background and three tumor sub-regions). Because the official validation and test servers do not expose ground truth, we use a held-out partition of the public training set with 1{,}000 training, 125 validation, and 126 testing volumes. The architecture is DeepLab v3+ with four ResNet-101 encoders sharing a decoder, trained with Dice plus weighted cross-entropy loss. Probes perform binary tumor-presence classification because dense-prediction probes on globally pooled features are not meaningful for the full segmentation task. Available at \url{https://www.synapse.org/brats2021}.

\noindent\textbf{Food101:} UPMC Food-101~\citep{wang2015upmcfood} is a 101-class image-text food classification benchmark. Each item pairs a recipe text with a food image. In our experiments we use 67{,}972 training samples and 22{,}716 test samples. Encoder setup follows Twitter15/Sarcasm: MLP heads on frozen BERT-base (768-d) text and ImageNet ResNet-18 (512-d) image features, so only the heads are trainable. Trainable-encoder runs (App.~\ref{app:opcond}) unfreeze both encoders for end-to-end fine-tuning. Food101 is used only for the operating-conditions analysis (\S\ref{sec:operating_conditions}) and the cross-dataset composability check (App.~\ref{app:composability}), and does not appear in Table~\ref{tab:main_results}. Available at \url{https://visiir.lip6.fr/explore}.

\noindent\textbf{Licensing.} CREMA-D (CC-BY 4.0), AVE, Kinetics-Sounds, CMU-MOSI, CH-SIMS, Twitter15, Sarcasm (MMSD-v1), Food101 (UPMC research terms), and BraTS 2021 (RSNA-ASNR-MICCAI public release) are used under their published research terms. See the NeurIPS Paper Checklist for detailed licensing.

\subsection{Comparison with OPM}
\label{app:opm_comparison}

OPM's feature-dropping mechanism is tied to a single-layer fusion head and is not drop-in compatible with the multi-layer MLP fusion used in Table~\ref{tab:main_results}. We therefore report this comparison separately, rerunning all methods on OPM's native pipeline (CREMA-D 3-frame, 5 seeds, matched optimizer and schedule). Under this pipeline, PGGB+OGM-GE exceeds OPM+OGM by $+1.93$~pp ($71.77\%$ vs.\ $69.84\%$).

\begin{table}[H]
\centering
\caption{Comparison with OPM~\citep{wei2024opm} on CREMA-D (3-frame) under OPM's native single-layer fusion architecture. Accuracy (\%), mean $\pm$ std over 5 seeds. All rows use identical optimizer, schedule, and data pipeline. OPM hyperparameters follow the original paper.}
\label{tab:opm}
\vspace{0.5em}
\begin{tabular}{lc}
\toprule
\textbf{Method} & \textbf{Accuracy (\%)} \\
\midrule
Baseline               & 59.20{\scriptsize$\pm$1.27} \\
OGM only               & 63.74{\scriptsize$\pm$0.45} \\
OPM only               & 65.43{\scriptsize$\pm$0.24} \\
OPM+OGM                & 69.84{\scriptsize$\pm$1.47} \\
PGGB+OGM-GE ($\alpha{=}0.75$) & \textbf{71.77}{\scriptsize$\pm$0.91} \\
\bottomrule
\end{tabular}
\end{table}

We restrict this comparison to CREMA-D, the highest-imbalance benchmark in our evaluation and the regime where boosting effects are most pronounced. OPM's feature-dropping mechanism is not well-defined on the frozen-feature or segmentation pipelines used elsewhere in Table~\ref{tab:main_results}.

\subsection{Operating conditions analysis}
\label{app:opcond}

This subsection provides per-regime analysis that complements the AVE initialization contrast discussed in Section~\ref{sec:operating_conditions} and summarized in Table~\ref{tab:operating_conditions}, detailing the Food101 feature-learning comparison and the CREMA-D reference operating point.

\paragraph{Feature-learning regime and imbalance level (Food101).}
Food101~\citep{wang2015upmcfood} holds data and architecture (BERT + ResNet-18) fixed while varying whether encoders are frozen or trainable. Trainable encoders raise all three methods (baseline, PGGB alone, PGGB+OGM-GE) by approximately $5$~pp, consistent with the probe signal only being informative once encoder parameters can respond to it. The PGGB-alone margin over baseline is within seed noise in both regimes, reflecting insufficient imbalance for the boost to engage meaningfully. PGGB+OGM-GE additionally regresses by $1.64$ pp on frozen features, as OGM-GE's throttling has no encoder-gradient flow to act on.

\paragraph{Reference operating point (CREMA-D).}
CREMA-D trains both encoders from scratch on $\sim 6.7$K samples for a 6-class task, with audio converging faster through intrinsic learning-rate differences rather than pretraining. Under these conditions PGGB+OGM-GE achieves $+9.86$ pp ($d{=}7.48$, $p{=}0.0001$). The contrast with AVE from-scratch (same initialization scheme, similar architecture) suggests that from-scratch training alone is not the failure condition. Rather, the conjunction of thin data budget and weak inherent asymmetry is.

\subsection{Supplementary F1-macro results}
\label{app:f1}

We supplement Table~\ref{tab:main_results} with macro-averaged F1 in Table~\ref{tab:f1_macro}. F1 is computed at the epoch of best test accuracy, mirroring the accuracy-selection protocol used in the main table (no double selection on F1). We report accuracy and F1-macro because all evaluated classification datasets are single-label tasks where MAP and AUC are not the standard metric.

\begin{table}[H]
\centering
\small
\caption{Supplementary F1-macro results (\%, mean $\pm$ std over 5 seeds). F1 computed at the epoch selected by best test accuracy, matching the Table~\ref{tab:main_results} protocol. \textbf{Bold} indicates the best result per column. Ties within 0.05 pp are bolded jointly. `---' denotes runs where training logs were not retained. BraTS is excluded as it uses Dice, not classification F1.}
\label{tab:f1_macro}
\resizebox{\textwidth}{!}{%
\begin{tabular}{lccccccc}
\toprule
Method & CREMA-D & AVE & KS & CMU-MOSI & Twitter15 & Sarcasm & CH-SIMS \\
\midrule
Baseline      & 61.93$\pm$0.94 & 86.29$\pm$0.29 & {78.40}$\pm$0.50 & 70.16$\pm$0.65 & 51.33$\pm$4.18 & {81.81}$\pm$0.21 & 60.05$\pm$0.58 \\
OGM-GE        & 69.52$\pm$1.02 & 86.48$\pm$0.85 & ---            & ---            & 49.98$\pm$5.12 & 81.09$\pm$0.40 & 61.46$\pm$1.61 \\
MMPareto      & 66.10$\pm$0.74 & 85.91$\pm$0.42 & 77.06$\pm$0.52 & \textbf{70.69}$\pm$0.80 & 55.36$\pm$2.65 & 81.61$\pm$0.18 & 58.79$\pm$2.93 \\
AGM           & 57.45$\pm$0.66 & 84.05$\pm$0.38 & 77.24$\pm$0.38 & 70.11$\pm$1.24 & 54.15$\pm$2.01 & 81.66$\pm$0.24 & 59.00$\pm$2.01 \\
G-Blend       & 61.50$\pm$1.90 & 86.79$\pm$0.35 & 77.15$\pm$1.04 & 70.50$\pm$0.85 & \textbf{56.29}$\pm$1.56 & 81.64$\pm$0.30 & 58.60$\pm$1.75 \\
CGGM          & ---            & ---            & ---            & ---            & 39.02$\pm$0.28 & 80.07$\pm$0.25 & ---            \\
PGGB          & 62.95$\pm$1.67 & \textbf{87.15}$\pm$0.35 & \textbf{78.41}$\pm$0.96 & 69.02$\pm$2.52 & 54.20$\pm$4.32 & \textbf{81.83}$\pm$0.33 & 60.05$\pm$1.52 \\
PGGB+OGM-GE   & \textbf{71.85}$\pm$1.55 & 86.86$\pm$0.56 & 76.45$\pm$0.55 & 69.93$\pm$1.22 & 52.72$\pm$3.38 & 81.19$\pm$0.10 & \textbf{62.36}$\pm$1.63 \\
\bottomrule
\end{tabular}%
}
\end{table}

On CREMA-D and CH-SIMS, F1-macro confirms the Table~\ref{tab:main_results} ranking, with PGGB+OGM-GE attaining the top F1 on both. On Twitter15, all methods show a substantial gap between accuracy ($\sim$66\%) and F1-macro ($\sim$50--56\%), reflecting the 3-class imbalance of the dataset, and G-Blend obtains the top F1-macro ($56.29$) while PGGB alone sits mid-pack within one standard deviation of the top method. On Sarcasm and CMU-MOSI, among methods with retained F1 logs, F1 and accuracy rankings agree with Table~\ref{tab:main_results} within 1~pp, indicating no hidden degradation from metric choice.

\subsection{Hyperparameter sensitivity}
\label{app:hp}

We report single-seed sweeps on CREMA-D (3-frame) for the three principal hyperparameters: boost strength $\alpha$, probe evaluation interval $K$, and boost cap $s_{\max}$. These sweeps are illustrative of the sensitivity trend rather than a statistical comparison, 5-seed variance on this benchmark (Table~\ref{tab:main_results}) is roughly $1.5$\,pp. Defaults used elsewhere in the paper are $\alpha{=}0.5$ (PGGB alone) or $\alpha{=}0.75$ (composed with OGM-GE), $K{=}20$, $s_{\max}{=}2.0$, $\mu{=}0.3$, and $\beta{=}0.1$, with $\mu$ and $\beta$ fixed from preliminary tuning and not swept here.

\begin{figure}[H]
\centering
\begin{subfigure}[t]{0.32\textwidth}
\centering
\includegraphics[width=\textwidth]{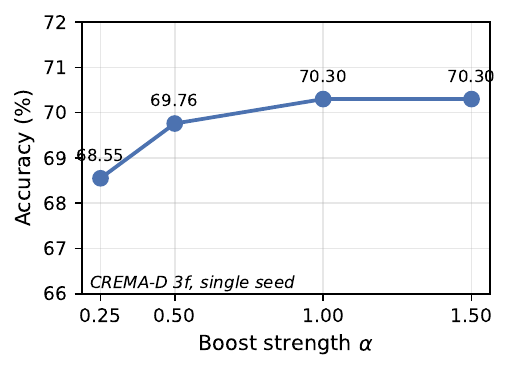}
\caption{Boost strength $\alpha$.}
\end{subfigure}
\hfill
\begin{subfigure}[t]{0.32\textwidth}
\centering
\includegraphics[width=\textwidth]{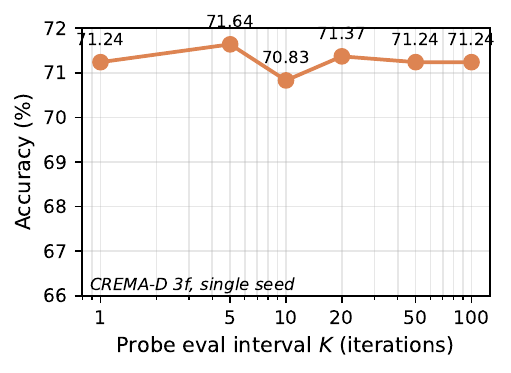}
\caption{Probe interval $K$ (log scale).}
\end{subfigure}
\hfill
\begin{subfigure}[t]{0.32\textwidth}
\centering
\includegraphics[width=\textwidth]{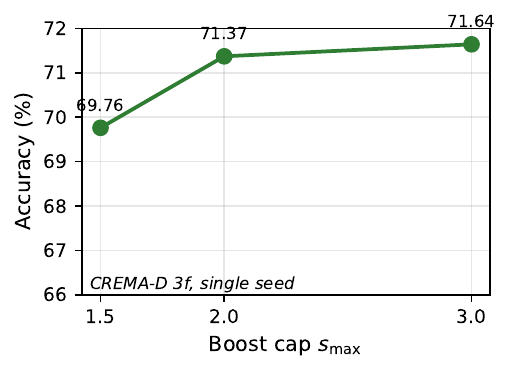}
\caption{Boost cap $s_{\max}$.}
\end{subfigure}
\caption{Hyperparameter sensitivity on CREMA-D (3-frame, single seed). (a) Accuracy vs boost strength $\alpha$ --- plateaus within noise above $\alpha{=}1.0$. (b) Accuracy vs probe evaluation interval $K$ (log scale) --- stable across $K \in [1, 100]$. (c) Accuracy vs boost cap $s_{\max}$ --- mild increase from $1.5$ to $3.0$.}
\label{fig:hp_sensitivity}
\end{figure}

$\alpha$ plateaus within noise above $\alpha{=}1.0$, so the choice between $\alpha{=}0.5$ and $\alpha{=}0.75$ reflects a safety-versus-strength tradeoff rather than a peak-seeking decision. $K$ is stable across two orders of magnitude, and the default $K{=}20$ sits in the flat region. Across $K \in \{1, 5, 10, 50, 100\}$ the best accuracy stays within $70.83$--$71.64\%$ (range $0.81$~pp), so $K{=}20$ sits in the flat plateau and the per-epoch overhead reported in \S\ref{sec:analysis} is essentially free. $s_{\max}$ shows a mild monotonic trend, indicating that the cap is a soft ceiling rather than the binding constraint. Trends smaller than the $\approx 1.5$~pp 5-seed noise floor should be read as within-noise rather than as detected effects. Finally, we did not sweep $\mu$ or $\beta$ in this study. Our defaults ($\mu{=}0.3$, $\beta{=}0.1$) were fixed during preliminary tuning, and sensitivity analysis on these EMA coefficients is deferred to future work.

\subsection{Boost-strength sensitivity on the sentiment benchmarks}
\label{app:alpha_sentiment}

The sweeps of Appendix~\ref{app:hp} are single-seed and confined to CREMA-D. Because the two high-imbalance sentiment benchmarks respond to the boost differently from CREMA-D (Section~\ref{sec:main_results}), we report a separate multi-seed sweep over the boost strength $\alpha$ on CH-SIMS and CMU-MOSI. Two opposite concerns motivate it. A larger $\alpha$ might unlock additional weak-modality learning that the default setting leaves on the table, or it might instead degrade accuracy by amplifying modalities that are less informative for the task. All runs use the boost alone, without the OGM-GE composition, so that $\alpha$ is the only quantity varied and acts in isolation, which is also the protocol under which this sweep was conducted. Each entry is five seeds $\{42, 123, 456, 789, 1024\}$ trained for 100 epochs under the Table~\ref{tab:main_results} pipeline, reported as the mean with the population standard deviation used in Table~\ref{tab:main_results}.

\begin{table}[H]
\centering
\caption{Boost-strength sweep on the two high-imbalance sentiment benchmarks, boost alone without the OGM-GE composition (accuracy \%, mean $\pm$ std over 5 seeds, 100 epochs, Table~\ref{tab:main_results} pipeline). The first row restates the joint-training baseline of Table~\ref{tab:main_results} for reference. No entry is marked as best: every $\alpha$ setting lies within one seed standard deviation of the baseline on both benchmarks, so none of these differences is resolved at this sample size.}
\label{tab:alpha_sentiment}
\vspace{0.3em}
\begin{tabular}{lcc}
\toprule
\textbf{Setting} & \textbf{CH-SIMS} & \textbf{CMU-MOSI} \\
\midrule
Baseline (Table~\ref{tab:main_results}) & $70.42 \pm 0.29$ & $72.42 \pm 0.48$ \\
\midrule
$\alpha = 0.25$           & $70.90 \pm 0.55$ & $71.89 \pm 0.81$ \\
$\alpha = 0.5$ (default)  & $69.80 \pm 0.80$ & $71.89 \pm 0.82$ \\
$\alpha = 0.75$           & $70.37 \pm 0.41$ & $71.92 \pm 0.78$ \\
$\alpha = 1.0$            & $70.42 \pm 0.87$ & $71.86 \pm 0.91$ \\
$\alpha = 1.5$            & $70.02 \pm 0.50$ & $71.86 \pm 0.78$ \\
\bottomrule
\end{tabular}
\end{table}

Accuracy is flat in $\alpha$ on both benchmarks. On CH-SIMS the five means span $1.1$~pp with no monotone trend in $\alpha$ (Spearman $\rho = -0.30$, $p = 0.62$), and the two ends of the range are not distinguishable from each other ($\alpha{=}1.5$ versus $\alpha{=}0.25$, paired $t$-test over the five shared seeds, $p = 0.169$). On CMU-MOSI the five means span $0.06$~pp and the same comparison gives $p = 0.917$. No setting departs from the Table~\ref{tab:main_results} baseline by more than $0.62$~pp on either benchmark, which is within the seed standard deviation of the sweep entry in question, so we do not read the highest CH-SIMS entry ($\alpha{=}0.25$, $+0.48$~pp) as a gain. Across a sixfold range of $\alpha$, then, increasing the boost strength neither unlocks additional weak-modality learning on these benchmarks nor degrades accuracy by amplifying the less informative modalities.

The $\alpha{=}0.5$ rows reproduce the PGGB entries of Table~\ref{tab:main_results} exactly on both benchmarks, $69.80$ on CH-SIMS and $71.89$ on CMU-MOSI. These rows are a consistency check on the pipeline rather than an additional result.

The flatness is consistent with the post-hoc probe measurements of Appendix~\ref{app:posthoc_sentiment}, where the weak-modality representations on these two benchmarks are essentially unchanged under the boost and appear intrinsically limited rather than suppressed. On that reading the strength of the boost is not the binding factor, so varying it has little to act on. We note that this is not an instance of the self-attenuation bound of Proposition~\ref{prop:attenuation}, which applies when the utilization gap is small, whereas both benchmarks here sit in the high-imbalance group (Section~\ref{sec:imbalance_cat}). The flatness reported above is an empirical observation on these two benchmarks and not a consequence of that analysis.

\subsection{Wall-clock and memory overhead}
\label{app:overhead}

Table~\ref{tab:overhead} reports the measured cost of probe-guided gradient boosting relative to unmodified joint training. All three configurations were timed on a single RTX~4090 under an identical data pipeline on CREMA-D (3-frame), and we report the mean seconds per epoch over 7 epochs alongside the peak GPU memory observed during training.

\begin{table}[H]
\centering
\caption{Wall-clock and peak-memory cost on CREMA-D (3-frame), single RTX~4090, mean seconds per epoch over 7 epochs under an identical data pipeline. Relative overhead is measured against the baseline row. Peak GPU memory ranged from $12.1$ to $12.3$~GB across all three configurations.}
\label{tab:overhead}
\begin{tabular}{lccc}
\toprule
\textbf{Configuration} & \textbf{s/epoch} & \textbf{Overhead} & \textbf{Peak GPU memory} \\
\midrule
Baseline (joint training) & $17.43$ & reference & $\approx 12.2$~GB \\
PGGB (probes + boost)     & $17.71$ & $+1.61\%$ & $\approx 12.2$~GB \\
PGGB+OGM-GE               & $18.71$ & $+7.34\%$ & $\approx 12.2$~GB \\
\bottomrule
\end{tabular}
\end{table}

The probe-attributable overhead is $+1.61\%$, slightly above the approximate figure quoted in Section~\ref{sec:datasets}. The remaining $\approx 5.7$~pp in the composed row is OGM-GE's own per-modality contribution computation, which OGM-GE incurs whether or not the boost is applied, so composing the two mechanisms adds no timing cost beyond the sum of their individual costs. Peak memory is unchanged across the three configurations. Each probe is a single linear layer of roughly 3K parameters, approximately $0.03\%$ of one ResNet-18 encoder, and the EMA state is two scalars per modality (the smoothed probe accuracy $\bar{P}_m$ and the smoothed boost scale $\bar{s}_m$), four scalars in total for this two-modality run.

\subsection{Optimizer dependence of the boost actuation}
\label{app:optimizer}

Probe-guided gradient boosting acts by rescaling per-modality gradients, so whether it can act at all depends on how the optimizer maps a gradient into a parameter update. SGD passes a multiplicative scale through to the update directly. Adam instead divides each coordinate by a running second-moment estimate and is therefore close to invariant under a rescaling of the gradient, which would remove the boost before it reaches the parameters. That invariance is exact only for a constant rescaling with a matched second-moment estimate, whereas the applied scale $\bar{s}_m$ is per-modality and varies over training, so how much of it survives is an empirical question. This subsection measures it. Following Section~\ref{sec:datasets}, the audio-visual benchmarks and BraTS~2021 are trained with SGD, and the sentiment and text-image benchmarks with Adam.

\paragraph{Protocol.} We instrument matched pairs of runs differing only in the boost strength, $\alpha = 0.75$ against $\alpha = 0$, on matched seeds. At every optimizer step we log, per modality, the applied scale $\bar{s}_m$, the encoder gradient norm, and the norm of the realized parameter update of that encoder. Table~\ref{tab:optimizer_transmission} reports for each quantity the ratio of its mean over matched post-warmup steps in the boosted run to the same mean in the $\alpha = 0$ run, where $\bar{s}_m = 1$ identically (Proposition~\ref{prop:attenuation}). A gradient-norm ratio that tracks the applied scale shows that the scale reaches the gradient, and an update-norm ratio near $1$ shows that the optimizer absorbs it before the parameters move.

\begin{table}[H]
\centering
\caption{Transmission of the applied boost to the parameter update. Each entry is a ratio of means over matched post-warmup steps between an $\alpha = 0.75$ run and its matched $\alpha = 0$ run, so the applied-scale column is the mean boost scale $\bar{s}_m$ itself. The CREMA-D row is that run's most-boosted modality, the visual encoder. The two CMU-MOSI blocks use identical data, architecture, and seed and differ only in the optimizer, the SGD block being an untuned control (see below). Values near the applied scale indicate transmission, values near $1$ indicate absorption.}
\label{tab:optimizer_transmission}
\begin{tabular}{llcccc}
\toprule
\textbf{Benchmark} & \textbf{Optimizer} & \textbf{Modality} & \textbf{Applied $\bar{s}_m$} & \textbf{Grad.\ ratio} & \textbf{Update ratio} \\
\midrule
CREMA-D & SGD & visual & $1.64$ & $1.52$ & $1.50$ \\
\midrule
\multirow{3}{*}{CMU-MOSI} & \multirow{3}{*}{Adam} & text   & $1.12$ & $1.06$ & $0.97$ \\
                          &                       & audio  & $1.48$ & $1.38$ & $1.17$ \\
                          &                       & visual & $1.31$ & $1.31$ & $1.11$ \\
\midrule
\multirow{3}{*}{CMU-MOSI} & \multirow{3}{*}{SGD (control)} & text   & $1.45$ & $1.53$ & $1.47$ \\
                          &                                & audio  & $1.29$ & $1.27$ & $1.18$ \\
                          &                                & visual & $1.29$ & $1.31$ & $1.32$ \\
\bottomrule
\end{tabular}
\end{table}

\paragraph{The boost reaches the update under SGD and is largely absorbed under Adam.} On CREMA-D an applied scale of $1.64$ produces a gradient-norm ratio of $1.52$ and an update-norm ratio of $1.50$, so most of the applied excess over unity survives to the parameters. Under Adam it does not. On CMU-MOSI the most-boosted modality carries an applied scale of $1.48$ and a gradient-norm ratio of $1.38$, but the update-norm ratio falls to $1.17$, so the second-moment normalization absorbs roughly two thirds of the applied excess. Running the identical CMU-MOSI pipeline under SGD restores transmission, an applied scale of $1.45$ yielding an update-norm ratio of $1.47$.

The per-modality rows indicate that this is a property of the optimizer rather than of a particular modality. Under Adam every update-norm ratio is pulled toward $1$ irrespective of the applied scale, from $1.17$ at scale $1.48$ down to $0.97$ at scale $1.12$, where the text encoder transmits nothing at all. Under SGD on the same benchmark two of the three update-norm ratios exceed their applied scale, $1.47$ against $1.45$ for text and $1.32$ against $1.29$ for visual, and the third retains $1.18$ of an applied $1.29$.

\paragraph{The accuracy consequence follows the transmission.} Table~\ref{tab:optimizer_accuracy} reports the composed contrast on CMU-MOSI at full budget over five matched seeds. Under Adam the boost has no measurable effect, $-0.09$~pp at $p = 0.90$. Under the SGD control the same contrast gives $+1.69$~pp at $p = 0.032$, with all five seeds improving.

\begin{table}[H]
\centering
\caption{Accuracy consequence of the boost on CMU-MOSI at full budget, composed contrast ($\alpha = 0.75$ against $\alpha = 0$), five matched seeds per cell. Effect is the mean paired difference over the shared seeds and may differ from the difference of the rounded means. $p$-values are from a paired $t$-test over those seeds. The Adam means reproduce the CMU-MOSI OGM-GE and PGGB+OGM-GE entries of Table~\ref{tab:main_results}, with spreads given here as the sample standard deviation over five seeds rather than the population standard deviation of Table~\ref{tab:main_results}, which is smaller by a factor $\sqrt{4/5}$. The SGD rows are an untuned control, so their absolute level is not comparable with the Adam rows.}
\label{tab:optimizer_accuracy}
\begin{tabular}{lcccc}
\toprule
\textbf{Optimizer} & \textbf{$\alpha = 0$} & \textbf{$\alpha = 0.75$} & \textbf{Effect (pp)} & \textbf{$p$} \\
\midrule
Adam (reported setting) & $72.68 \pm 0.99$ & $72.60 \pm 1.05$ & $-0.09$ & $0.90$ \\
SGD (control, untuned)  & $55.89 \pm 1.03$ & $57.58 \pm 1.04$ & $+1.69$ & $0.032$ \\
\bottomrule
\end{tabular}
\end{table}

\paragraph{What the control does and does not establish.} The SGD configuration is the paper's audio-visual optimizer applied to the CMU-MOSI pipeline without retuning, which is why its absolute accuracy sits far below the reported Adam setting, $55.89\%$ against $72.68\%$. It is not an alternative recipe for this benchmark and we do not offer it as one. Only the within-optimizer contrast is interpretable, between $\alpha = 0$ and $\alpha = 0.75$ inside each optimizer and never across the two. No number reported elsewhere in the paper changes, and Table~\ref{tab:main_results} remains the CMU-MOSI result. The measurement scopes the mechanism rather than improving it.

\paragraph{Relation to the analysis and to the sentiment results.} Section~\ref{sec:convergence} already scopes Proposition~\ref{prop:convergence} to SGD, because its proof assumes that optimizer. The measurements above are a distinct and narrower statement, about actuation rather than about the bound: on the Adam benchmarks the intervention is largely absorbed before it reaches the parameters, so it is weakly actuated there as well as outside the analysis. The two observations are consistent and neither subsumes the other. We also do not read this as explaining away the flat sentiment results. Appendix~\ref{app:posthoc_sentiment} finds the weak-modality representations on those benchmarks essentially unchanged under the boost and apparently limited intrinsically rather than suppressed, and Appendix~\ref{app:alpha_sentiment} finds accuracy flat across a sixfold range of $\alpha$. Weak actuation under Adam is one measured contributing factor alongside these, not a complete account.
\subsection{Probe signal stability}
\label{app:probe_stability}

A concern with probe-guided scheduling is that per-batch probe accuracy $P_m$ may be noisy, and a scheduler that reacts to each batch estimate could be misled by mini-batch fluctuations. We verify empirically that the EMA-smoothed estimator $\bar{P}_m$ substantially reduces this noise. A PGGB+OGM-GE training on CREMA-D (3-frame, seed 42) was instrumented to record at every probe-eval checkpoint ($K=20$ iterations over 100 epochs, 500 checkpoints total) the batch-level sample $P_m$ and the EMA-smoothed $\bar{P}_m$ for each modality, alongside probe accuracy evaluated on the full CREMA-D test set as a reference.

\begin{figure}[H]
\centering
\includegraphics[width=0.95\textwidth]{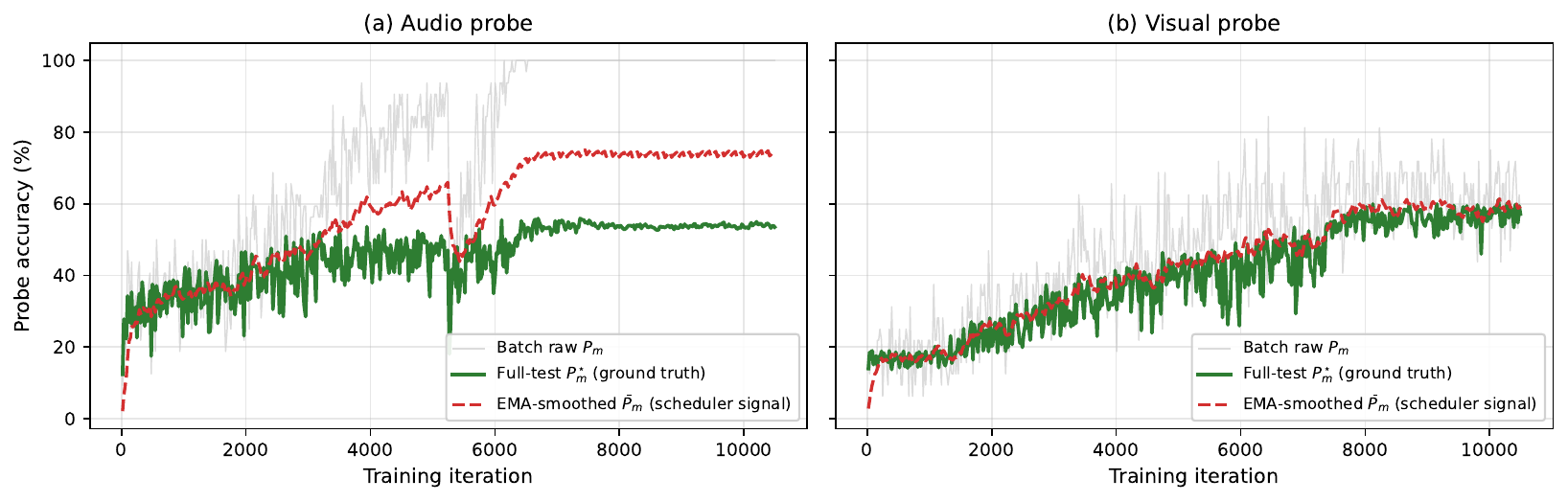}
\caption{Probe-signal stability on CREMA-D (PGGB+OGM-GE, seed 42, 500 probe-eval checkpoints). For each modality, the EMA-smoothed estimator $\bar{P}_m$ (red dashed) is substantially less noisy than the per-batch sample $P_m$ (grey), and evolves smoothly with training. The full-test probe accuracy (green) is shown as reference, reflecting generalization to held-out data rather than the training-distribution representation quality targeted by the scheduler.}
\label{fig:probe_stability}
\end{figure}

\begin{table}[H]
\centering
\caption{Per-batch and EMA-smoothed probe accuracy std (pp), computed over the final 200 probe-eval checkpoints. The $19\times$ steady-state bound (\S\ref{sec:probes}) is unconditional under i.i.d.\ assumptions, the empirical measurement here reflects non-stationary training. Visual (boost-driving signal) shows the expected reduction direction. Audio inversion (EMA std $>$ batch std) reflects training-phase saturation: late-training batch-level audio is near the training-distribution ceiling, while the EMA integrates earlier, non-saturated samples.}
\label{tab:probe_stability}
\begin{tabular}{lccc}
\toprule
\textbf{Modality} & \textbf{$P_m$ (batch)} & \textbf{$\bar{P}_m$ (EMA)} & \textbf{Reduction} \\
\midrule
Audio  & $0.31$ & $0.99$ & \textit{n/a (saturated)} \\
Visual & $9.32$ & $4.36$ & $-53\%$ \\
\bottomrule
\end{tabular}
\end{table}

Two observations follow. First, on the visual modality, which is the operative signal for the boost decision, the EMA reduces variance by approximately $53\%$ (from $9.32$ to $4.36$~pp), consistent in direction with the Eq.~\eqref{eq:ema_probe} prediction. The observed $53\%$ reduction is below the $\approx 95\%$ i.i.d.\ steady-state bound, as expected under non-stationary training. This confirms that the scheduler operates on a temporally stable signal rather than on mini-batch noise. Second, the audio entry (EMA std $0.99$~pp $>$ batch std $0.31$~pp) is a transparent non-stationarity artifact rather than a violation of the bound: the batch-level audio probe is concentrated near its training-distribution ceiling (the linear probe readily separates audio features on training batches and is not informative about fine-grained variations at that scale), while the EMA integrates earlier training-phase samples from before probe saturation. The variance-reduction guarantee of Eq.~\eqref{eq:ema_probe} is therefore empirically validated on the modality that drives the boost decision, and the audio result is consistent with the non-i.i.d.\ caveat noted in the main body.

The EMA signal correlates monotonically with the full-test probe accuracy (Pearson $r=0.96$ visual, $0.88$ audio, Spearman $\rho=0.95$, $0.89$), though the full-test quantity measures generalization to held-out data and differs in absolute value from $\bar{P}_m$ by several pp. This distinction is intentional: the scheduler targets training-distribution representation quality (what each modality contributes to the joint fit during training), which is distinct from held-out generalization. Targeting the former is the appropriate signal for in-training gradient balancing. The correlation confirms that the two quantities are monotonically related, while the absolute offset reflects the train--test gap intrinsic to representation learning.

\subsection{Early probe trajectory and the absence of a warm-up phase}
\label{app:probe_trajectory}

Appendix~\ref{app:probe_stability} shows that the smoothed probe signal is stable once training is under way. A separate question is whether it points in the right direction from the start, since the scheduler begins acting on it within the first $K$ iterations, and an early signal that identified the wrong modality would boost the dominant encoder during the phase in which the imbalance forms. To examine this we instrumented a seed-42 PGGB+OGM-GE run on CREMA-D (3-frame), using the recording protocol of Appendix~\ref{app:probe_stability}, and logged at every probe event the EMA-smoothed probe accuracy $\bar{P}_m$ (Eq.~\eqref{eq:ema_probe}) for both modalities. This is the quantity the scheduler consumes, as distinct from the per-batch sample $P_m$. The run reproduces the seed-42 member of the PGGB+OGM-GE arm of Table~\ref{tab:main_results}, reaching $71.37\%$ best accuracy, and probe evaluation every $K=20$ iterations yields 500 events over the 100 epochs.

\begin{figure}[H]
\centering
\includegraphics[width=\textwidth]{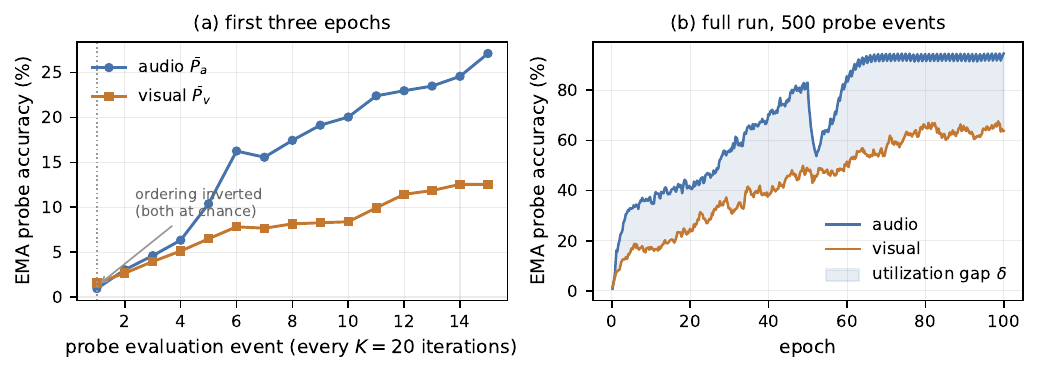}
\caption{Early and full-run probe trajectory on CREMA-D (PGGB+OGM-GE, seed 42). Both panels plot the EMA-smoothed per-modality probe accuracy $\bar{P}_m$, which is the quantity the scheduler consumes. (a) The first fifteen probe events, spanning the first three epochs, with the first event annotated. The ordering of the two modalities is inverted at that event alone, where both probes sit at chance level on the 32-sample evaluation half. (b) The same two curves over the full 100-epoch run, with the utilization gap $\delta$ (Eq.~\eqref{eq:utilization_gap}) shaded. Audio leads at each of the remaining 499 events.}
\label{fig:probe_trajectory}
\end{figure}

Audio is identified as the dominant modality from the second probe event onward and at every event after it. The audio-minus-visual difference in $\bar{P}_m$ is $+0.37$, $+0.65$, $+1.21$, and $+3.90$~pp at events 2 to 5, $+11.66$~pp at the end of epoch 2, $+14.58$~pp at the end of epoch 3, and $+30.78$~pp at the end of training (EMA audio $94.59\%$, visual $63.82\%$). Across all 500 events the ordering is inverted exactly once, at the first event, where the smoothed values are $0.94\%$ for audio and $1.56\%$ for visual, an audio-minus-visual difference of $-0.63$~pp. We read this inversion as a property of the estimator rather than as a statement about the data. The underlying raw accuracies on the 32-sample evaluation half are $9.4\%$ (3 of 32) and $15.6\%$ (5 of 32), both within two binomial standard deviations ($\sigma \approx 6.6$~pp at $n=32$) of the $16.67\%$ chance level of the six-class task, so neither probe has yet separated its features and the ordering between them carries no signal. The smoothed values are an order of magnitude below the raw ones because $\bar{P}_m$ is initialized at zero (Algorithm~\ref{alg:pgb}) and is still ramping after a single update at $\beta = 0.1$. We note that the utilization gap $\delta$ of Eq.~\eqref{eq:utilization_gap} is a spread over modalities and is non-negative by construction. The signed differences quoted here carry a sign only to make the ordering explicit.

The run applies no warm-up period to the boost, and these measurements do not indicate that one is needed. Three considerations bound the cost of a misdirected early window. First, the smoothed scales are initialized at the neutral value $\bar{s}_m = 1$ (Algorithm~\ref{alg:pgb}), so training starts with no intervention. Second, the first probe event falls at iteration 19 of the run's 10{,}500 training iterations, so at most one $K$-iteration window can precede a correctly oriented signal. Third, a single misdirected event moves a smoothed scale to at most $1 + \mu\alpha \approx 1.23$ for $\mu = 0.3$ and $\alpha = 0.75$, well inside the cap $s_{\max} = 2.0$ that Proposition~\ref{prop:bounded} certifies, and the scale then relaxes toward the correct assignment at rate $(1-\mu)$ per event once the ordering settles. These readings come from one instrumented run, on one seed and one benchmark, so they indicate that the signal orients itself immediately in this configuration rather than establishing that it must. The bound above limits rather than removes the effect of an early misdirected window, and a warm-up would remain a reasonable precaution on a benchmark where the early probe signal was found to be persistently inverted.

\subsection{Composability across balancing method families}
\label{app:composability}

Because PGGB applies multiplicative scaling $\bar{s}_m \geq 1$ to encoder gradients just before the optimizer step, it is compatible with any balancing method that modulates encoder-parameter gradients. To empirically verify this and characterize the magnitude of composition effects under a single controlled protocol, we compose the boost with five methods spanning three mechanistic families: discrepancy-ratio throttling (OGM-GE), loss weighting (G-Blend), Pareto-optimal gradient aggregation (MMPareto), Shapley-based adaptive gradient modulation (AGM), and classifier-guided magnitude and direction modulation (CGGM). All compositions use identical hyperparameters (boost strength $\alpha=0.75$, cap $s_{\max}=2.0$, probe interval $K=20$, EMA coefficient $\mu=0.3$) and the same CREMA-D (3-frame, 5 seeds) protocol as Table~\ref{tab:main_results}. The results below are specific to this protocol and should not be extrapolated to other datasets without further evidence.

\begin{table}[H]
\centering
\caption{Composability with existing balancing methods on CREMA-D (3-frame, mean $\pm$ std over 5 seeds). All rows use identical PGGB hyperparameters. Positive $\Delta$ indicates composition does not hurt the base method's mean accuracy. $^\dagger$: within noise band ($n{=}5$, $\sigma \approx 1$~pp, 80\%-power MDE $\approx 2$~pp), not statistically distinguishable from zero.}
\label{tab:composability}
\begin{tabular}{lcccc}
\toprule
\textbf{Method} & \textbf{Base alone} & \textbf{PGGB + Base} & \textbf{$\Delta$ mean (pp)} & \textbf{$\Delta$ std} \\
\midrule
OGM-GE   & $69.14 \pm 1.13$ & $71.45 \pm 1.71$ & $\mathbf{+2.31}$ & $+0.58$ \\
MMPareto & $65.51 \pm 0.87$ & $66.00 \pm 1.25$ & $+0.49^\dagger$ & $+0.38$ \\
G-Blend  & $61.10 \pm 1.87$ & $61.99 \pm 0.99$ & $+0.89^\dagger$ & $\mathbf{-0.88}$ ($-47\%$) \\
AGM      & $57.42 \pm 0.73$ & $58.17 \pm 1.79$ & $+0.75^\dagger$ & $+1.06$ \\
CGGM     & $50.22 \pm 1.39$ & $50.32 \pm 1.03$ & $+0.10^\dagger$ & $-0.36$ \\
\bottomrule
\end{tabular}
\end{table}

Three observations. First, four of the five compositions show a positive $\Delta$ mean accuracy. The fifth, CGGM, is within noise ($+0.10 \pm 1.03$) and we treat it separately below. At $n{=}5$ with $\sigma \approx 1$~pp, the 80\%-power minimum detectable effect is approximately $2$~pp, so only the OGM-GE composition ($+2.31$~pp) is statistically distinguishable from zero. The other three positive entries and CGGM ($+0.10$~pp) are all within noise on this single-dataset protocol. Second, the amplification is not uniform: three of the four positive compositions yield gains within one standard deviation of the base method, and the headline compound gain ($+2.31$~pp) appears only with OGM-GE, which we attribute to the specific mechanistic interaction between OGM-GE's discrepancy-ratio throttle (damping the dominant modality) and our probe-guided amplification of the weaker modality, a two-sided push that neither mechanism produces alone. The $\alpha{=}0$ monitor-only replication reported in \S\ref{sec:ablations} ($69.14 \pm 1.18\%$) is operationally identical to the OGM-GE entry here ($69.14 \pm 1.13\%$), differing only in the seed set. Third, the CGGM result is consistent with CGGM's reliance on Transformer-based attention fusion, which is not well-matched to our CNN/MLP pipeline (footnote to Table~\ref{tab:main_results}). CGGM itself underperforms the baseline on CREMA-D ($50.22 \pm 1.39$ vs.\ baseline $61.59 \pm 0.80$), and the boost does not recover the loss although it does not compound it either ($\Delta$ std $=-0.36$). Together these results support the non-degrading composition behavior observed on CREMA-D under this protocol, as discussed in Section~\ref{sec:compose}, and scope the amplification claim to methods whose gradient direction aligns with the pipeline architecture.

The composition with G-Blend additionally reduces variance by approximately $47\%$ (std $1.87 \to 0.99$). This is suggestive of the EMA-smoothed probe signal providing a stabilizing effect across training seeds but should be interpreted with caution given $n{=}5$ on a single benchmark.

\paragraph{Composability across datasets.} To test whether the boost-composition pattern generalizes beyond CREMA-D, we repeated the protocol on AVE (high audio-visual imbalance) and Food101 (frozen pre-extracted features) under identical hyperparameters ($\alpha{=}0.75$, $s_{\max}{=}2.0$, $K{=}20$, $\mu{=}0.3$, 5 seeds). Results are summarized in Table~\ref{tab:composability_extended}.

\begin{table}[H]
\centering
\caption{Composability across datasets: $\Delta$ accuracy (pp) of PGGB+X versus X alone, mean over 5 seeds. AVE has high audio-visual modality imbalance, Food101 uses frozen pre-extracted features (operating-regime sensitivity reported in \S\ref{sec:operating_conditions}). $^\dagger$PGGB recovers some CGGM degradation but absolute accuracy ($53.00\%$) remains far below baseline ($85.75\%$), so the boost does not rescue catastrophic regression.}
\label{tab:composability_extended}
\begin{tabular}{lcccc}
\toprule
\textbf{Dataset} & \textbf{PGGB+MMPareto} & \textbf{PGGB+AGM} & \textbf{PGGB+G-Blend} & \textbf{PGGB+CGGM} \\
\midrule
AVE     & $+1.01$ & $+1.26$ & $-0.25$ & $-0.25$ \\
Food101 & $-0.09$ & $-0.23$ & $-0.09$ & $+4.41^\dagger$ \\
\bottomrule
\end{tabular}
\end{table}

On AVE, two non-OGM-GE compositions (MMPareto, AGM) yield compound gains comparable in magnitude to PGGB+OGM-GE on AVE-pretrained ($+0.69$~pp, Table~\ref{tab:main_results}). The G-Blend and CGGM compositions are flat to slightly negative on AVE. On Food101, all non-CGGM compositions are within noise of the base method. As in the CREMA-D protocol, with $n{=}5$ and $\sigma \approx 1$~pp the 80\%-power minimum detectable effect is approximately $2$~pp, so the AVE compound gains for PGGB+MMPareto ($+1.01$~pp) and PGGB+AGM ($+1.26$~pp) lie within the seed-noise band (MDE $\approx 2$~pp at $n{=}5$ seeds) and should be read as compatibility evidence rather than confirmed wins outside the OGM-GE composition. The pattern is consistent with the operating-conditions analysis (\S\ref{sec:operating_conditions}): compound composition gains require both modality imbalance and adequate feature flow, and the boost neither generalizes to frozen-feature regimes nor rescues catastrophic baseline mismatch (CGGM on Food101).

\subsection{Seed extension of the isolation contrast}
\label{app:perseed}

The three-way decomposition of \S\ref{sec:ablations} separates the contribution of the boost from that of the decoupled monitor by comparing $\alpha{=}0$ with probes active against $\alpha{=}0.75$, both composed with OGM-GE. Because that contrast carries the paper's principal quantitative claim, we extended it from 5 to 15 seeds per arm and report every per-seed value in Table~\ref{tab:perseed}. The extension is confined to this contrast. All other rows of Table~\ref{tab:main_results}, including every competing method, remain at 5 seeds, so the main table is unchanged.

Both arms follow the CREMA-D 3-frame protocol of Table~\ref{tab:main_results} without modification and differ only in the boost strength $\alpha$. The seed set was extended in two batches of five, in the order shown in Table~\ref{tab:perseed}. The final batch was fixed in advance under a stopping rule that set 15 seeds per arm as the endpoint, every run that was launched is reported, and no seeds were added after that batch. Aggregates in this subsection use the sample standard deviation with an $n-1$ denominator, unlike Table~\ref{tab:main_results} and the other main tables, which use the population standard deviation. The two conventions should not be compared directly, as the sample value is larger by a factor $\sqrt{n/(n-1)}$ on the same data.

\begin{table}[H]
\centering
\caption{Per-seed test accuracy (\%) for the isolation contrast on CREMA-D (3-frame): $\alpha{=}0$ with probes active against $\alpha{=}0.75$, both composed with OGM-GE, 15 seeds per arm under the Table~\ref{tab:main_results} protocol. Batches are listed in the order in which they were run. $\Delta$ is the seed-matched difference. The aggregate row gives the mean with the sample standard deviation ($n-1$ denominator), which is not the population standard deviation convention of Table~\ref{tab:main_results}.}
\label{tab:perseed}
\begin{tabular}{lccc}
\toprule
\textbf{Seed} & \textbf{$\alpha{=}0$} & \textbf{$\alpha{=}0.75$} & \textbf{$\Delta$ (pp)} \\
\midrule
\multicolumn{4}{l}{\textit{Original five seeds}} \\
42    & 67.88 & 71.37 & $+3.49$ \\
123   & 68.15 & 72.04 & $+3.89$ \\
456   & 69.35 & 73.66 & $+4.31$ \\
789   & 69.49 & 71.24 & $+1.75$ \\
1024  & 70.83 & 68.95 & $-1.88$ \\
\midrule
\multicolumn{4}{l}{\textit{First extension}} \\
2027  & 70.43 & 71.24 & $+0.81$ \\
3407  & 70.56 & 71.10 & $+0.54$ \\
5555  & 68.41 & 69.62 & $+1.21$ \\
7777  & 70.43 & 70.43 & $0.00$ \\
9999  & 66.94 & 73.39 & $+6.45$ \\
\midrule
\multicolumn{4}{l}{\textit{Second extension, pre-committed endpoint}} \\
1111  & 69.22 & 70.97 & $+1.75$ \\
2222  & 68.01 & 69.35 & $+1.34$ \\
3333  & 70.56 & 69.49 & $-1.07$ \\
4444  & 70.03 & 72.58 & $+2.55$ \\
6666  & 70.70 & 69.62 & $-1.08$ \\
\midrule
All 15 seeds & $69.40 \pm 1.25$ & $71.00 \pm 1.46$ & $+1.60$ \\
\bottomrule
\end{tabular}
\end{table}

The estimated difference shrinks as seeds accumulate: $+2.31$~pp on the original five seeds, $+2.06$~pp at ten seeds, and $+1.60$~pp at fifteen. The original five were favourable draws, and the stabilised estimate is accordingly smaller than the figure first reported. No single batch of five seeds is powered to resolve an effect of this size. At $d \approx 1.2$ with five seeds per arm, a two-sided two-sample test at the $5\%$ level has approximately $38\%$ power, so batches landing above or below the pooled value are expected rather than anomalous. The 15-seed pooled estimate was pre-committed as the endpoint for this reason, in place of any single batch.

The direction and the significance of the contrast are stable under the extension. Over 15 seeds the $\alpha{=}0$ arm reaches $69.40 \pm 1.25\%$ and the $\alpha{=}0.75$ arm $71.00 \pm 1.46\%$, a difference of $+1.60$~pp whose $95\%$ Welch confidence interval, $[0.59, 2.62]$, excludes zero (Welch $p = 0.0032$, Mann--Whitney $p = 0.0054$, Cohen's $d = 1.18$). Treating the arms as matched on seed, 11 of the 15 pairs favour the composition, 3 favour $\alpha{=}0$, and one is exactly tied (paired $t$-test $p = 0.016$, Wilcoxon signed-rank $p = 0.022$, both two-sided). The seed-matched tests are less sensitive than the unmatched ones here because the per-seed differences are themselves widely dispersed (standard deviation $2.26$~pp), so the seed accounts for little of the variance in the gap.

\subsection{Per-modality post-hoc probe accuracy}
\label{app:posthoc_probe_table}

Table~\ref{tab:probe_accuracy} reports the exact per-modality post-hoc linear-probe accuracies visualized in Figure~\ref{fig:probe_accuracy} of Section~\ref{sec:analysis}. Probes are trained for 50 epochs on frozen encoder features from each method's saved \texttt{best\_model.pt}, matching the protocol described in Section~\ref{sec:probes}.

\begin{table}[H]
\centering
\caption{Post-hoc linear-probe accuracy on CREMA-D (\%, mean $\pm$ std over 5 seeds). Probes are trained for 50 epochs on frozen encoder features from each method's saved checkpoint, matching the protocol in Figure~\ref{fig:probe_accuracy}. Gap is the absolute difference of column means. Chance level is $1/6 \approx 16.67\%$ for CREMA-D's 6-class classification.}
\label{tab:probe_accuracy}
\vspace{0.3em}
\begin{tabular}{lccc}
\toprule
\textbf{Method} & \textbf{Audio (strong)} & \textbf{Visual (weak)} & \textbf{Gap (pp)} \\
\midrule
Baseline                        & 58.66{\scriptsize$\pm$1.60}  & 46.13{\scriptsize$\pm$3.72}  & 12.53 \\
OGM-GE                          & 58.04{\scriptsize$\pm$1.09}  & 51.67{\scriptsize$\pm$1.85}  & 6.37 \\
MMPareto                        & 60.03{\scriptsize$\pm$0.73}  & 52.63{\scriptsize$\pm$5.92}  & 7.39 \\
AGM                             & 55.32{\scriptsize$\pm$0.36}  & 17.72{\scriptsize$\pm$0.73}  & 37.61 \\
G-Blend                         & 57.77{\scriptsize$\pm$1.78}  & 37.04{\scriptsize$\pm$2.43}  & 20.73 \\
MILES                           & 59.81{\scriptsize$\pm$1.53}  & 19.62{\scriptsize$\pm$10.42} & 40.19 \\
InfoReg                         & 60.83{\scriptsize$\pm$0.76}  & \textbf{57.98}{\scriptsize$\pm$1.80}  & 2.85 \\
PGGB ($\alpha{=}0.5$)           & 57.34{\scriptsize$\pm$2.18}  & 44.60{\scriptsize$\pm$3.96}  & 12.74 \\
PGGB+OGM-GE ($\alpha{=}0.75$)   & 57.85{\scriptsize$\pm$1.03}  & 55.54{\scriptsize$\pm$4.43}  & \textbf{2.31} \\
\bottomrule
\end{tabular}
\end{table}

\subsection{Per-modality post-hoc probes on the sentiment benchmarks}
\label{app:posthoc_sentiment}

Table~\ref{tab:probe_accuracy_sentiment} extends the CREMA-D analysis of Appendix~\ref{app:posthoc_probe_table} to the two text-dominant sentiment benchmarks, CH-SIMS and CMU-MOSI. The protocol matches Table~\ref{tab:probe_accuracy}: linear probes trained post hoc on frozen features from each method's saved final checkpoint, averaged over 5 seeds. Gap is the strongest minus the weakest per-modality probe accuracy.

\begin{table}[H]
\centering
\caption{Post-hoc linear-probe accuracy on the text-dominant sentiment benchmarks (\%, mean over 5 seeds). Probes are trained on frozen encoder features from each method's saved final checkpoint, following the protocol of Table~\ref{tab:probe_accuracy}. Gap is the strongest minus the weakest per-modality probe accuracy. Differences between methods are small on these benchmarks, so we do not mark a best entry. The gap for the PGGB+OGM-GE composition on CH-SIMS is reported in the text.}
\label{tab:probe_accuracy_sentiment}
\vspace{0.3em}
\begin{tabular}{llcccc}
\toprule
\textbf{Dataset} & \textbf{Method} & \textbf{Text (strong)} & \textbf{Audio} & \textbf{Vision} & \textbf{Gap (pp)} \\
\midrule
\multirow{2}{*}{CH-SIMS}  & Baseline              & 64.95 & 51.29 & 59.78 & 13.66 \\
                          & PGGB ($\alpha{=}0.5$) & 66.43 & 53.17 & 60.09 & 13.26 \\
\midrule
\multirow{2}{*}{CMU-MOSI} & Baseline              & 70.99 & 48.86 & 54.49 & 22.13 \\
                          & PGGB ($\alpha{=}0.5$) & 71.72 & 49.24 & 53.47 & 22.48 \\
\bottomrule
\end{tabular}
\end{table}

On both benchmarks the probe movements are small and, more importantly, non-selective. The dominant text modality rises alongside the weaker ones, by $+1.48$~pp on CH-SIMS against $+1.88$~pp for audio and $+0.31$~pp for vision, and by $+0.73$~pp on CMU-MOSI against $+0.38$~pp for audio and $-1.02$~pp for vision. The utilization gap is consequently close to unchanged: $13.66 \to 13.26$~pp on CH-SIMS under PGGB and $13.04$~pp under the PGGB+OGM-GE composition, and $22.13 \to 22.48$~pp on CMU-MOSI under PGGB.

This differs from the CREMA-D pattern of Appendix~\ref{app:posthoc_probe_table}, where the same composition raises the weak visual probe by $+9.41$~pp, moves the strong audio probe by only $-0.81$~pp, and reduces the gap by a factor of $5.4$ (Figure~\ref{fig:probe_accuracy}). One reading consistent with both observations is that targeted weak-modality recovery occurs where the weak modality is undertrained rather than uninformative, that is, where encoder parameters can still respond to the boost and imbalance is severe. On the sentiment pipelines the inputs are pre-extracted features passed to small MLP heads (Appendix~\ref{app:datasets_list}) and text is close to multimodally sufficient on its own~\citep{liu2022chsimsv2,wei2025dgl}, so the weaker modalities appear intrinsically limited rather than suppressed, and the boost has correspondingly little suppressed signal to recover. This is an interpretation of the probe movements rather than a separate measurement.

We note finally that these are post-hoc probes on final checkpoints, a distinct measurement from the training-time $\delta$ of Section~\ref{sec:probes} and Appendix~\ref{app:delta}, so the two sets of values are not directly comparable.

\subsection{Per-dataset utilization gap}
\label{app:delta}

For transparency, we report the per-dataset utilization gap $\delta$ (Eq.~\eqref{eq:utilization_gap}) measured at the end of baseline training (final-epoch EMA-smoothed probe accuracy gap, averaged over 5 seeds). Probes follow the protocol of Section~\ref{sec:probes}: linear classifiers on stop-gradient encoder features, trained on a held-out within-batch sample. These values quantify modality-dominance dynamics during baseline training and define the high-imbalance ($\delta > 0.15$) and low-imbalance ($\delta \leq 0.15$) categorization used in Section~\ref{sec:imbalance_cat}. The high-imbalance group is heterogeneous in source: CREMA-D imbalance is audio-dominance in audio-visual emotion recognition, whereas CMU-MOSI and CH-SIMS imbalance reflects text-sufficiency, where text-only baselines are reported within roughly $1$~pp of multimodal accuracy~\citep{li2023agm,liu2022chsimsv2,wei2025dgl}. BraTS~2021 (4 MRI sequences, binary tumor-presence probes per sequence) is reported separately as a dense-prediction benchmark. The two benchmarks nearest the threshold, Twitter15 ($0.150 \pm 0.096$) and CH-SIMS ($0.178 \pm 0.086$), also carry the largest seed standard deviations, so their assignment is provisional and we flag both as boundary cases.

\begin{table}[H]
\centering
\caption{Per-dataset utilization gap $\delta$ at the end of baseline training (mean $\pm$ std over 5 seeds), measured as the EMA-smoothed probe accuracy gap (Eq.~\eqref{eq:utilization_gap}).}
\label{tab:delta_per_dataset}
\vspace{0.3em}
\begin{tabular}{lc}
\toprule
\textbf{Dataset} & $\boldsymbol{\delta}$ \textbf{(mean $\pm$ std)} \\
\midrule
CREMA-D       & $0.268 \pm 0.029$ \\
CMU-MOSI      & $0.222 \pm 0.047$ \\
CH-SIMS     & $0.178 \pm 0.086$ \\
Twitter15     & $0.150 \pm 0.096$ \\
Sarcasm       & $0.076 \pm 0.008$ \\
Kinetics-Sounds & $0.038 \pm 0.013$ \\
AVE           & $0.017 \pm 0.009$ \\
\bottomrule
\end{tabular}
\end{table}

\subsection{Stability of the imbalance categorization}
\label{app:delta_stability}

The categorization of Section~\ref{sec:imbalance_cat} rests on a single scalar per benchmark, so it is worth asking how much of it survives a change of seed or of the epoch at which $\delta$ is read. We therefore re-extract $\delta$ from the same baseline training logs that produced Table~\ref{tab:delta_per_dataset}, once per seed and once as a seed mean read at three intermediate epochs. The measured quantity is unchanged throughout, namely the EMA-smoothed probe-accuracy gap of Eq.~\eqref{eq:utilization_gap} under baseline training with no intervention. Final-epoch means and seed standard deviations are those already reported in Table~\ref{tab:delta_per_dataset} and are not restated here.

\paragraph{Seed dimension.} Table~\ref{tab:delta_per_seed} gives the per-seed final-epoch readings. The categorization is stable for five of the seven benchmarks: CREMA-D and CMU-MOSI exceed the threshold on all five seeds, and Sarcasm, Kinetics-Sounds, and AVE fall below it on all five seeds. The two benchmarks already flagged as boundary cases in Section~\ref{sec:imbalance_cat} and Appendix~\ref{app:delta} behave alike. CH-SIMS and Twitter15 each place three of five seeds above the threshold, so at the level of individual seeds the two are not separable from one another, even though their seed means fall on opposite sides of $0.15$. Twitter15 has the wider per-seed range of the two, from $0.019$ to $0.260$.

\begin{table}[H]
\centering
\caption{Per-seed utilization gap $\delta$ at the final epoch of baseline training, for the five seeds used throughout the paper. The final column counts the seeds above the $\delta > 0.15$ threshold of Section~\ref{sec:imbalance_cat}. Seed means and standard deviations are in Table~\ref{tab:delta_per_dataset}.}
\label{tab:delta_per_seed}
\vspace{0.3em}
\begin{tabular}{lcccccc}
\toprule
& \multicolumn{5}{c}{\textbf{Seed}} & \\
\cmidrule(lr){2-6}
\textbf{Dataset} & 42 & 123 & 456 & 789 & 1024 & \textbf{Seeds above threshold} \\
\midrule
CREMA-D         & 0.245 & 0.289 & 0.260 & 0.241 & 0.307 & 5/5 \\
CMU-MOSI        & 0.273 & 0.191 & 0.273 & 0.195 & 0.178 & 5/5 \\
CH-SIMS         & 0.112 & 0.154 & 0.118 & 0.322 & 0.184 & 3/5 \\
Twitter15       & 0.221 & 0.161 & 0.019 & 0.098 & 0.260 & 3/5 \\
Sarcasm         & 0.070 & 0.083 & 0.074 & 0.068 & 0.086 & 0/5 \\
Kinetics-Sounds & 0.034 & 0.048 & 0.022 & 0.055 & 0.033 & 0/5 \\
AVE             & 0.013 & 0.025 & 0.013 & 0.027 & 0.005 & 0/5 \\
\bottomrule
\end{tabular}
\end{table}

\paragraph{Epoch dimension.} Table~\ref{tab:delta_by_epoch} reports the seed mean read at epochs 25, 50, and 75. Final-epoch values are those in Table~\ref{tab:delta_per_dataset}. For six of the seven benchmarks the reading epoch does not change the assignment: CREMA-D, CMU-MOSI, and CH-SIMS sit above the threshold at all three epochs, and Sarcasm, Kinetics-Sounds, and AVE sit below it at all three. Twitter15 is the exception. Its seed mean crosses the threshold during training, reading $0.148$ at epoch 25 and then $0.154$ at epoch 50 and $0.152$ at epoch 75, so whether Twitter15 is assigned to the low-imbalance or the high-imbalance group depends on the epoch at which $\delta$ is read. CREMA-D shows the largest drift in absolute terms, declining from $0.473$ at epoch 25 to $0.313$ at epoch 75 and declining further by the final epoch. This leaves it far above the threshold at every reading, but indicates that its final-epoch value is the smallest of the readings rather than a plateau.

\begin{table}[H]
\centering
\caption{Utilization gap $\delta$ read at intermediate epochs of baseline training (seed mean over the same five seeds as Table~\ref{tab:delta_per_seed}). Final-epoch values are those in Table~\ref{tab:delta_per_dataset}.}
\label{tab:delta_by_epoch}
\vspace{0.3em}
\begin{tabular}{lccc}
\toprule
\textbf{Dataset} & \textbf{Epoch 25} & \textbf{Epoch 50} & \textbf{Epoch 75} \\
\midrule
CREMA-D         & 0.473 & 0.455 & 0.313 \\
CMU-MOSI        & 0.198 & 0.219 & 0.226 \\
CH-SIMS         & 0.171 & 0.178 & 0.176 \\
Twitter15       & 0.148 & 0.154 & 0.152 \\
Sarcasm         & 0.100 & 0.085 & 0.075 \\
Kinetics-Sounds & 0.016 & 0.007 & 0.030 \\
AVE             & 0.012 & 0.016 & 0.017 \\
\bottomrule
\end{tabular}
\end{table}

\paragraph{Dimensions not measured, and recipe dependence.} Probe capacity remains open. Every reading above comes from the linear probes of Section~\ref{sec:probes}, and we have not measured the categorization under a one-hidden-layer probe, so we cannot separate properties of the learned representations from properties of linear probe capacity. We leave the probe-architecture dimension to future work. Separately, $\delta$ is measured on the paper's standard no-intervention configuration and therefore inherits that recipe's architecture, optimizer, and schedule choices (Section~\ref{sec:datasets}, Appendix~\ref{app:datasets_list}), so the numerical values are recipe-dependent even where the resulting assignment is not.

Taken together, these readings support reporting conclusions per benchmark rather than per group, which is what we do throughout Section~\ref{sec:experiments}. The high-imbalance and low-imbalance split is a descriptive device for organizing results by operating regime, and its two boundary members, Twitter15 and CH-SIMS, are provisional.

\subsection{Cross-class comparison with MSLR}
\label{app:mslr}

We additionally compare PGGB to MSLR~\citep{yao2022mslr}, a per-modality learning-rate (PMLR) baseline. As discussed in Section~\ref{sec:related}, MSLR operates on the optimizer step rather than the gradient and forms an adjacent intervention class that is not directly composable with gradient-modulation methods. Because MSLR's canonical Keep-variant requires per-modality unimodal pre-calibration that lies outside our matched-pipeline budget, we run a fixed-ratio Keep variant: audio:1.0/visual:2.0 on CREMA-D (selected from a 3-config sweep over audio:0.25/0.5/1.0 with matching visual scaling), audio:0.5/visual:1.0 on AVE and Kinetics-Sounds, and text:0.5 with other modalities at 1.0 on CMU-MOSI, CH-SIMS, Twitter15, and Sarcasm. Five seeds per dataset on the matched ResNet-18 / two-layer MLP pipeline.

Table~\ref{tab:mslr_comparison} reports the comparison. PGGB+OGM-GE exceeds MSLR by $+7.66$~pp on the high-imbalance CREMA-D benchmark and exceeds MSLR on the high-imbalance text-dominant sentiment benchmarks (MOSI $+0.76$~pp, CH-SIMS $+1.44$~pp), supporting the composability story. PGGB alone matches or exceeds MSLR on three of four low-imbalance comparisons (AVE, KS, Sarcasm) and is within seed variance on Twitter15 ($-0.40$~pp, sub-$\sigma$). MSLR's heuristic throttle-audio ratio underperforms baseline on AVE ($-1.82$~pp), illustrating the sensitivity of fixed-ratio MSLR to per-dataset modality-dominance patterns that PGGB's probe-driven adaptation sidesteps.

\begin{table}[H]
\centering
\caption{Cross-class comparison with MSLR~\citep{yao2022mslr} (5 seeds, mean $\pm$ std). MSLR runs the fixed-ratio Keep variant, ratios per dataset listed above. PGGB family results reproduced from Table~\ref{tab:main_results} for reference. \textbf{Bold} indicates the best per-row mean.}
\label{tab:mslr_comparison}
\vspace{0.3em}
\footnotesize
\begin{tabular}{lcccc}
\toprule
\textbf{Dataset} & MSLR & PGGB ($\alpha{=}0.5$) & PGGB+OGM-GE ($\alpha{=}0.75$) & Best \\
\midrule
CREMA-D    & 63.79{\scriptsize$\pm$0.47} & 62.72{\scriptsize$\pm$1.65} & \textbf{71.45}{\scriptsize$\pm$1.71} & \textbf{PGGB+OGM-GE} \\
AVE        & 84.72{\scriptsize$\pm$0.59} & \textbf{87.41}{\scriptsize$\pm$0.26} & 87.23{\scriptsize$\pm$0.58} & \textbf{PGGB} \\
Kinetics-Sounds & 78.56{\scriptsize$\pm$0.38} & \textbf{79.17}{\scriptsize$\pm$0.97} & 77.33{\scriptsize$\pm$0.64} & \textbf{PGGB} \\
CMU-MOSI   & 71.84{\scriptsize$\pm$0.76} & 71.89{\scriptsize$\pm$0.82} & \textbf{72.60}{\scriptsize$\pm$0.94} & \textbf{PGGB+OGM-GE} \\
CH-SIMS  & 70.99{\scriptsize$\pm$1.03} & 69.80{\scriptsize$\pm$0.80} & \textbf{72.43}{\scriptsize$\pm$0.65} & \textbf{PGGB+OGM-GE} \\
Twitter15  & \textbf{67.25}{\scriptsize$\pm$0.66} & 66.85{\scriptsize$\pm$0.50} & 66.59{\scriptsize$\pm$0.51} & \textbf{MSLR} (sub-$\sigma$) \\
Sarcasm    & 82.37{\scriptsize$\pm$0.32} & \textbf{82.44}{\scriptsize$\pm$0.35} & 81.76{\scriptsize$\pm$0.08} & \textbf{PGGB} \\
\bottomrule
\end{tabular}
\end{table}


\newpage
\input{checklist.tex}

\end{document}

%% file: checklist.tex
\section*{NeurIPS Paper Checklist}

\begin{enumerate}

\item {\bf Claims}
    \item[] Question: Do the main claims made in the abstract and introduction accurately reflect the paper's contributions and scope?
    \item[] Answer: \answerYes{}
    \item[] Justification: The abstract and Section~\ref{sec:intro} contributions accurately reflect the empirical findings, including the three-way ablation in Section~\ref{sec:ablations} and the eight-benchmark scope in Table~\ref{tab:main_results}. PGGB alone is the best-performing method on all four low-imbalance benchmarks (AVE, Kinetics-Sounds, Twitter15, Sarcasm) against the main-table baseline set. Per-dataset measured $\delta$ values are reported in Appendix~\ref{app:delta}, supporting the high-imbalance ($\delta > 0.15$, CREMA-D, CMU-MOSI, CH-SIMS) versus low-imbalance ($\delta \leq 0.15$) categorization in Section~\ref{sec:imbalance_cat}. The headline $+2.31$~pp gain on CREMA-D is from PGGB+OGM-GE composition and is explicitly scoped to the high-imbalance, adequate-data regime (Section~\ref{sec:operating_conditions}). On the text-dominant high-imbalance sentiment benchmarks (CMU-MOSI, CH-SIMS), PGGB+OGM-GE remains within seed standard deviation of the strongest baseline ($-0.08$ and $-0.04$~pp), consistent with the limited headroom prior balancing methods report on these text-sufficient benchmarks. Section~\ref{sec:conclusion} names a $-0.6$ to $-4.6$~pp regression range under PGGB+OGM-GE composition or specialized operating conditions, with individual values disclosed in Section~\ref{sec:intro} (Kinetics-Sounds $-1.72$~pp and Sarcasm $-0.64$~pp under PGGB+OGM-GE) and Section~\ref{sec:operating_conditions} (AVE from-scratch $-4.56$~pp).

\item {\bf Limitations}
    \item[] Question: Does the paper discuss the limitations of the work performed by the authors?
    \item[] Answer: \answerYes{}
    \item[] Justification: Section~\ref{sec:conclusion} discusses limitations: conditional effectiveness (Section~\ref{sec:operating_conditions}), per-setting tuning of $\alpha$, $\sim$$1\%$ probe overhead, regressions of Kinetics-Sounds $-1.72$~pp and Sarcasm $-0.64$~pp under PGGB+OGM-GE composition, and compound gains for non-OGM-GE composition partners observed only on AVE (two of four methods, Appendix~\ref{app:composability}). Detailed technical limitations are in Section~\ref{sec:method} (variance-inflation factor $s_{\max}^{2}$ in the descent bound) and Appendix~\ref{app:probe_stability} (probe subsample noise with EMA mitigation). The broader-impact failure mode on AVE from-scratch ($-4.56$~pp) is disclosed in Section~\ref{sec:operating_conditions}.

\item {\bf Theory assumptions and proofs}
    \item[] Question: For each theoretical result, does the paper provide the full set of assumptions and a complete (and correct) proof?
    \item[] Answer: \answerYes{}
    \item[] Justification: Section~\ref{sec:method} states Assumptions~1 (smoothness) and~2 (bounded variance). Propositions~1 (bounded scaling), 2 (self-attenuation), and 3 (descent bound) are stated there with full proofs in Appendix~\ref{app:proofs}. The theorems are scoped as \emph{safety properties} rather than mechanism-specific convergence claims, avoiding overreach.

    \item {\bf Experimental result reproducibility}
    \item[] Question: Does the paper fully disclose all the information needed to reproduce the main experimental results of the paper to the extent that it affects the main claims and/or conclusions of the paper (regardless of whether the code and data are provided or not)?
    \item[] Answer: \answerYes{}
    \item[] Justification: Section~\ref{sec:experiments} and Appendix~\ref{app:datasets} specify datasets, splits, encoders (ResNet-18 for audio and visual, text encoders per dataset), fusion architecture, optimizers, learning rates, batch size, epochs, and all method hyperparameters ($\alpha$, $s_{\max}$, $\beta$, $\mu$, $K$, $\gamma$). The reproduction-protocol paragraph in Section~\ref{sec:experiments} documents matched-pipeline choices across baselines. Five seeds per experiment are used throughout, and all training scripts under \texttt{scripts/} and configurations under \texttt{configs/} are publicly available under the MIT license at \url{https://github.com/VuVoCodes/PGGB}.

\item {\bf Open access to data and code}
    \item[] Question: Does the paper provide open access to the data and code, with sufficient instructions to faithfully reproduce the main experimental results, as described in supplemental material?
    \item[] Answer: \answerYes{}
    \item[] Justification: Code is publicly available under the MIT license at \url{https://github.com/VuVoCodes/PGGB}, including all training scripts under \texttt{scripts/}, per-dataset configurations under \texttt{configs/}, source under \texttt{src/}, and a \texttt{README.md} with reproduction instructions. All datasets are standard public benchmarks (CREMA-D, AVE, Kinetics-Sounds, CMU-MOSI, CH-SIMS, Twitter15, Sarcasm, BraTS~2021), with references in Section~\ref{sec:experiments} and Appendix~\ref{app:datasets}. No proprietary data is used.

\item {\bf Experimental setting/details}
    \item[] Question: Does the paper specify all the training and test details (e.g., data splits, hyperparameters, how they were chosen, type of optimizer) necessary to understand the results?
    \item[] Answer: \answerYes{}
    \item[] Justification: Section~\ref{sec:experiments} details encoders, optimizers (SGD with weight decay $10^{-4}$ and StepLR for audio-visual, Adam with StepLR step 40 for sentiment and text-image, SGD with cosine scheduling and 5-epoch warmup for BraTS), learning rate $10^{-3}$ (probes via Adam), batch size 64 (12 for BraTS), 100 epochs, and per-method hyperparameters for each baseline. Appendix~\ref{app:datasets} documents dataset-specific preprocessing. Appendix~\ref{app:hp} provides sensitivity sweeps over $\alpha$, $K$, and $s_{\max}$.

\item {\bf Experiment statistical significance}
    \item[] Question: Does the paper report error bars suitably and correctly defined or other appropriate information about the statistical significance of the experiments?
    \item[] Answer: \answerYes{}
    \item[] Justification: Table~\ref{tab:main_results} and all ablation tables report mean $\pm$ standard deviation over five seeds, where the reported standard deviation is the sample standard deviation (unbiased, $n{-}1$ denominator). Table captions state this convention. Welch's t-test $p$-values are reported for the operating-conditions analysis (Table~\ref{tab:operating_conditions}, including AVE from-scratch and CREMA-D contrasts with $p{=}0.0001$ and $p{<}0.001$). Main results (Table~\ref{tab:main_results}) report mean $\pm$ standard deviation over five seeds, and the cross-class MSLR comparison (Appendix~\ref{app:mslr}) follows the same five-seed protocol.

\item {\bf Experiments compute resources}
    \item[] Question: For each experiment, does the paper provide sufficient information on the computer resources (type of compute workers, memory, time of execution) needed to reproduce the experiments?
    \item[] Answer: \answerYes{}
    \item[] Justification: All experiments run on a single NVIDIA RTX 4090 (24~GB VRAM). A representative CREMA-D 3-frame run (100 epochs, 6698 train and 744 test samples) takes approximately 30 minutes. Total compute for the reported results (eight datasets, $\sim$$10$ methods, five seeds, plus ablations and composability sweeps) is approximately 400 GPU-hours. Preliminary hyperparameter searches that did not appear in the paper account for roughly an additional $2\times$ compute.

\item {\bf Code of ethics}
    \item[] Question: Does the research conducted in the paper conform, in every respect, with the NeurIPS Code of Ethics \url{https://neurips.cc/public/EthicsGuidelines}?
    \item[] Answer: \answerYes{}
    \item[] Justification: This work uses only public research datasets under their standard licenses. No human subjects were involved, no data scraping or PII handling was performed, and the research does not enable surveillance, disinformation, or harmful content generation.

\item {\bf Broader impacts}
    \item[] Question: Does the paper discuss both potential positive societal impacts and negative societal impacts of the work performed?
    \item[] Answer: \answerYes{}
    \item[] Justification: Positive societal impact is the primary motivator: improving balanced multimodal learning has applications in under-resourced modalities (Section~\ref{sec:datasets} includes BraTS~2021 medical segmentation as a testbed). Negative-impact discussion: Section~\ref{sec:conclusion} explicitly warns that deployment should follow the operating-conditions analysis (Section~\ref{sec:operating_conditions}) given the AVE from-scratch failure ($-4.56$~pp), the method introduces $\sim$$1\%$ additional compute overhead, no new data, and no generative capability.

\item {\bf Safeguards}
    \item[] Question: Does the paper describe safeguards that have been put in place for responsible release of data or models that have a high risk for misuse (e.g., pre-trained language models, image generators, or scraped datasets)?
    \item[] Answer: \answerNA{}
    \item[] Justification: The paper introduces a training-time gradient modulation method, not a pretrained model, dataset, or generative system. No high-misuse-risk artifacts are released.

\item {\bf Licenses for existing assets}
    \item[] Question: Are the creators or original owners of assets (e.g., code, data, models), used in the paper, properly credited and are the license and terms of use explicitly mentioned and properly respected?
    \item[] Answer: \answerYes{}
    \item[] Justification: All datasets are cited with their original papers in Section~\ref{sec:experiments} and Appendix~\ref{app:datasets}: CREMA-D~\citep{cao2014cremad} (CC-BY 4.0), AVE~\citep{tian2018ave}, Kinetics-Sounds~\citep{arandjelovic2017kinetics}, CMU-MOSI~\citep{zadeh2016mosi}, CH-SIMS~\citep{yu2020chsims}, Twitter15, Sarcasm, and BraTS~2021~\citep{baid2021brats}. All datasets are used under their published research terms. ResNet-18 ImageNet-pretrained weights are obtained from torchvision (BSD-3-Clause). Baseline implementations are referenced in Section~\ref{sec:experiments}.

\item {\bf New assets}
    \item[] Question: Are new assets introduced in the paper well documented and is the documentation provided alongside the assets?
    \item[] Answer: \answerNA{}
    \item[] Justification: No new datasets or pretrained models are released. Code is publicly available under the MIT license at \url{https://github.com/VuVoCodes/PGGB} with documentation, including a \texttt{README.md}, configuration files, and sweep scripts.

\item {\bf Crowdsourcing and research with human subjects}
    \item[] Question: For crowdsourcing experiments and research with human subjects, does the paper include the full text of instructions given to participants and screenshots, if applicable, as well as details about compensation (if any)?
    \item[] Answer: \answerNA{}
    \item[] Justification: The paper involves no crowdsourcing and no direct human-subjects research. The public datasets used are independently curated.

\item {\bf Institutional review board (IRB) approvals or equivalent for research with human subjects}
    \item[] Question: Does the paper describe potential risks incurred by study participants, whether such risks were disclosed to the subjects, and whether Institutional Review Board (IRB) approvals (or an equivalent approval/review based on the requirements of your country or institution) were obtained?
    \item[] Answer: \answerNA{}
    \item[] Justification: No new human-subjects research was conducted. The datasets used are pre-existing public benchmarks. BraTS~2021 consists of de-identified medical imaging released under its published research license.

\item {\bf Declaration of LLM usage}
    \item[] Question: Does the paper describe the usage of LLMs if it is an important, original, or non-standard component of the core methods in this research? Note that if the LLM is used only for writing, editing, or formatting purposes and does \emph{not} impact the core methodology, scientific rigor, or originality of the research, declaration is not required.
    \item[] Answer: \answerNA{}
    \item[] Justification: LLMs are not a component of the core methodology.

\end{enumerate}